\documentclass[10pt,journal,compsoc]{IEEEtran}

\usepackage{graphicx}
\usepackage{balance}  
\usepackage{float}
\usepackage{supertabular}
\usepackage{caption}
\usepackage{url}
\usepackage{algorithm}
\usepackage{algpseudocode}
\usepackage{xcolor}
\usepackage{footmisc}
\usepackage{enumitem}
\usepackage{xspace}
\usepackage{amssymb, amsmath, amsthm}
\usepackage{multirow}

\usepackage{graphicx}
\usepackage{subfig}

\theoremstyle{definition}
\newtheorem{definition}{Definition}

\theoremstyle{problems}

\theoremstyle{example}
\newtheorem{example}{Example}

\theoremstyle{property}
\newtheorem{property}{Property}

\theoremstyle{lemma}

\ifCLASSOPTIONcompsoc
  \usepackage[nocompress]{cite}
\else
  \usepackage{cite}
\fi

\usepackage[normalem]{ulem}
\usepackage{soul}

\newcommand{\jianxin}[1]{\textcolor{red}{#1}}

\newcommand{\taige}[1]{\textcolor{violet}{#1}}

\begin{document}
\title{
Distributed Optimization of Graph Convolutional Network using Subgraph Variance
}

%
%
        
\author{
Taige Zhao, Xiangyu Song, Jianxin Li$^{*}$\thanks{* Corresponding author}, Wei Luo, Imran Razzak
\IEEEcompsocitemizethanks{
\IEEEcompsocthanksitem Taige Zhao, Xiangyu Song, Jianxin Li, Wei Luo, Imran Razzak are with the School of Information Technology, Deakin University, Geelong, VIC3125, Australia.\protect\\
E-mail: \{zhaochr, xiangyu.song, jianxin.li, wei.luo, imran.razzak\}@deakin.edu.au}
\thanks{Manuscript received April 19, 2005; revised August 26, 2015.}}

\markboth{IEEE Transactions on Parallel and Distributed Systems, ~Vol.~xx, No.~xx, October.~2021}%
{Zhao \MakeLowercase{\textit{et al.}}: Distributed Optimization of Graph Convolutional Network using Subgraph Variance}

\IEEEtitleabstractindextext{%
\begin{abstract}
In recent years, Graph Convolutional Networks (GCNs) have achieved great success in learning from graph-structured data. With the growing tendency of graph nodes and edges, GCN training by single processor cannot meet the demand for time and memory, which led to a boom into distributed GCN training frameworks research. However, existing distributed GCN training frameworks require enormous communication costs between processors since multitudes of dependent nodes and edges information need to be collected and transmitted for GCN training from other processors.
 To address this issue, we propose a \textbf{G}raph \textbf{A}ugmentation based \textbf{D}istributed GCN framework \textbf{(GAD)}. In particular, GAD has two main components, \textbf{\textit{GAD-Partition}} and \textbf{\textit{GAD-Optimizer}}. We first propose a graph augmentation-based partition \textbf{\textit{(GAD-Partition)}} that can divide original graph into augmented subgraphs to reduce communication by selecting and storing as few significant nodes of other processors as possible while guaranteeing the accuracy of the training. In addition, we further design a subgraph variance-based importance calculation formula and propose a novel weighted global consensus method, collectively referred to as \textbf{\textit{GAD-Optimizer}}. This optimizer adaptively reduces the importance of subgraphs with large variances for the purpose of reducing the effect of extra variance introduced by GAD-Partition on distributed GCN training. Extensive experiments on four large-scale real-world datasets demonstrate that our framework significantly reduces the communication overhead ($\approx50\%$), improves the convergence speed ($\approx2X$) of distributed GCN training, and slight gain in accuracy ($\approx0.45\%$) based on minimal redundancy compared to the state-of-the-art methods.

\end{abstract}

\begin{IEEEkeywords}
Graph convolutional network, Distribution optimize, Graph augmentation, Subgraph variance, Communicate reduction
\end{IEEEkeywords}}
\maketitle

\IEEEdisplaynontitleabstractindextext
\IEEEpeerreviewmaketitle

\section{Introduction}

\IEEEPARstart{I}{n} many real-world application scenarios, the underlying data is often modelled and represented as graph-structured data, such as social networks~\cite{DBLP:journals/complexity/ZhaoCCL20}, Internet of Things~\cite{DBLP:journals/tsp/RomeroMG17} and traffic networks\cite{DBLP:journals/access/AhmedPCAL21}. Graph Convolutional Networks (GCNs) have been widely used with remarkable success in recent years to learn representations of these graph-structured data. The power of GCNs in capturing the dependencies of graph via message passing between its nodes, enables it to get vector representations of nodes, edges or graphs, and improve the prediction accuracy on many downstream tasks such as node classification\cite{DBLP:conf/kdd/ChiangLSLBH19, DBLP:conf/iclr/ChenMX18, DBLP:conf/nips/Huang0RH18}, link prediction\cite{DBLP:conf/iclr/SatorrasE18}, and graph classification\cite{DBLP:journals/ijon/MaWZTA21}. This breakthrough has led us to apply GCNs to increasingly complex scenarios, normally composed of giant networks containing hundreds of millions of nodes and billions of edges. For example, Microsoft Academic Graph \cite{DBLP:conf/kdd/TangZYLZS08} contains 111 million nodes and approximately one billion edges and the knowledge graph of Freebase \cite{DBLP:conf/www/TanonVSSP16} contains nearly 2 billion nodes and almost 30 billion edges. Along with this comes a huge demand for time and memory space for GCN training. Specifically, the number of support nodes potentially grows exponentially with the number of GCN layers to update nodes embedding, making it computationally and space complex.

Recently, many attempts have been committed to alleviating the difficulties of memory and time overhead in GCN training through distributed implementations. However, it brings several new challenges, such as how to allocate the processor memory rationally, balance the load and optimize the efficiency of communication between processors \cite{DBLP:journals/tpds/ZhaoLCGJHQWWZLL22, DBLP:journals/tpds/ChungSRPGACK17}. 
Preliminary research including PinSage \cite{DBLP:conf/kdd/YingHCEHL18}, AliGraph \cite{DBLP:journals/pvldb/ZhuZYLZALZ19} and DGL \cite{DBLP:journals/corr/abs-1909-01315} targeted these core issues. AliGraph separates the traditional underlying architecture into storage, sampling, and operation layers and applies a caching mechanism to store intermediate results. PinSage uses the local convolution separation method to calculate the final convolution result for the whole graph dynamically. DGL integrates the information dissemination mechanism into the message and reduction function to transfer and aggregate information, enhancing the framework's scalability without adding additional overhead.



 However, most of these approaches still fail to guarantee efficiency and accuracy due to extensive communication load in distributed GCN training. 
 Jiang et al.\cite{DBLP:journals/corr/abs-2101-07706} and Scardapane et al.\cite{DBLP:journals/tsipn/ScardapaneSL21} emphasized on reduction of communication between GPUs or processors and presented neighbor sampling method. Specifically, Jiang et al.\cite{DBLP:journals/corr/abs-2101-07706} reduced the communication overhead by assigning the smaller probability to the remote node and higher sampling probabilities to the local nodes. Higher the probability at local nodes results more likely these nodes to be sampled and less likely the remote node, which results in the reduction of communication load. Similarly, Scardapane et al.\cite{DBLP:journals/tsipn/ScardapaneSL21} used an additional concatenation subgraph to record the shortest communication path and split the graph among different agents. 
 Even though these methods showed a significant reduction in communication overhead, however, still have communication overhead, especially for big graphs and the large number of processors. 
 Angerd et al.\cite{ DBLP:journals/corr/abs-2012-04930} alleviate the communication overhead issue by 
randomly selecting a fixed number of nodes from other processors and replicating these nodes in local processors. However, it needs to manually specify the number of replicated nodes and the number of hops for selecting the replicated nodes. Thus, it is challenging to achieve high accuracy, low replication, and low time cost for distributed training of GCN without manually determine these parameters. 

 To address the aforementioned challenges, 
 in this work, we propose a novel \underline{G}raph \underline{A}ugmentation based \underline{D}istributed GCN framework (GAD) with two main components, GAD-Partition and GAD-Optimizer. Inspired by Angerd et al.\cite{ DBLP:journals/corr/abs-2012-04930}, we first develop a graph augmentation-based partition (GAD-Partition) to divide a original graph into subgraphs with the addition of augmented information. Specifically, we use Metis\cite{DBLP:journals/jpdc/KarypisK98a} to split the graph, which can minimize the candidate replication nodes for each subgraphs. 
 And then, we devise a Monte-Carlo-based nodes importance measurement and a depth-first sampling method to replicate the significant nodes from candidate replication nodes for augmenting each subgraph without manually specifying the number of replicated nodes and the number of hops. 
Besides, we develop a subgraph variance-based optimizer (GAD-Optimizer) to accelerate Dist-GCN training and improve training accuracy. Since subgraph variance as a good metric can be used to measure the degree distribution of nodes and difference between nodes within a subgraph, i.e., the lower the variance of a subgraph, the closer the structural feature of its nodes in the subgraph could be.  
 Therefore, it is faster to engage the convergence of the GCN training propagation and training accuracy using a group of nodes with lower subgraph variance than that of higher subgraph variance \cite{DBLP:conf/icml/ChenZS18}. We extend the subgraph variance measurement formula to subgraphs generated by partition methods. We then provide a novel weighted global consensus mechanism to unify the GCN parameters in each processor and accelerate the training speed and accuracy by considering the effect of subgraph variance on the gradient.

The \textbf{contributions} of this work are summarized as follows:
\begin{enumerate}

\item[$\bullet$] We propose a novel Graph Augmentation-based Distributed GCN framework, namely GAD, to reduce the communication costs, accelerate the training time, and guarantee higher training accuracy in dealing with large-scale graph data.


\item[$\bullet$] An adaptive graph partition method, namely GAD-partition, is developed to partition a original graph into the augmented subgraphs. Based node importance measurement mechanism, it can reduce the communication costs of processors by adaptively selecting and storing as few significant replicate nodes as possible, while guaranteeing high training accuracy.


\item[$\bullet$] We design a subgraph variance-based importance measurement and devise a novel weighted global consensus strategy, collectively referred to as GAD-Optimizer. It can adaptively adjust the weight of each subgraph based on the corresponding importance to accelerate the convergence of the training. 



\item[$\bullet$] To demonstrate the higher performance of GAD, we implement six state-of-the-art distributed GCN training methods, and compare with our proposed GAD on four real-world graph datasets.


\end{enumerate}

The rest of this paper is organised as follows. Section 2 reviews the related work.
Section 3 describes the detailed procedure of GAD-Partition and GAD-Optimizer. 
Finally, We report the experimental setting and the evaluation results in Section 4, and conclude our work in Section 5.
\section{Related Work}
There is a substantial body of research related to minimization of overheads in computation and communication across multiple processors in graph neural networks. Below, we first survey distributed graph representation learning before discussing in-depth optimized distributed graph representation learning.

\subsection{Distributed graph representation learning} 
The aim of graph representation learning is to get low dimensional continuous representation from discrete graph and preserve the properties of graphs (i.e. node similarity) in embedding space \cite{DBLP:journals/kbs/LiWLZ21, chen2021effective, DBLP:journals/corr/abs-2103-15447}. In order to accelerate the training process and deal with large-scale graph data, several distributed GNN training frameworks such as PyTorch Geometric\cite{DBLP:journals/corr/abs-1903-02428}, DGL\cite{DBLP:journals/corr/abs-1909-01315}, AGL\cite{DBLP:journals/pvldb/ZhangHL0HSGWZ020} were developed recently. They perform graph training on multiple GPUs or distributed machines. 
Scardapan et al.\cite{DBLP:journals/tsipn/ScardapaneSL21} presented fully distributed graph convolutional network. It introduces a global consensus algorithm to unify the GCN model parameter by average gradient. However, these methods still need to take high frequent communication costs among the processors because boundary nodes need to access and aggregate their neighboring nodes' information from other processors during the model training process.


\subsection{Optimized distributed graph representation learning} To reduce processors' communication in a distributed environment, recent research focus on development of optimized distributed representation learning. 
Ali-Graph\cite{DBLP:journals/pvldb/ZhuZYLZALZ19} used  
a novel storage layer to cache the nodes and their intermediate result to reduce communication between local and other processors. 
Similarly, Bai et al.\cite{DBLP:journals/tpds/BaiLLWMLX21} presented an efficient data loader to store frequently accessed nodes in cache by using a novel indexing algorithm which results in speeding up the acquisition of information between processors to reduce communication time. Jiang et al.\cite{DBLP:journals/corr/abs-2101-07706} provided different sampling probability for nodes on current processor and other processors. By assigning a higher sampling probability to the local nodes, it can reduce the frequency of remote nodes being visited and achieve the purpose of reducing communication costs. Zhao et al.\cite{DBLP:journals/corr/abs-2011-09430} 
designed an efficient scheduling transmission network to find the minimal processor communication route for reducing communication time. 
To further reduce the communication between processors,
a line of research proposes to replicate nodes across processors. For instance, DistDGL\cite{DBLP:journals/corr/abs-2010-05337} develops 
a strategy to reduce the communication by random replicate nodes and edges to be communicated through a given probability from other processors. Angerd et al.\cite{DBLP:journals/corr/abs-2012-04930} 
proposed an sampler to augment each mini-batch with a small amount of edges and nodes from all other sub-graphs uniformly and randomly.
Although these distributed optimization methods based on the sampling strategy can avoid node and edge communication between processors, the redundancy in dense graph is too large to be acceptable, due to the unexpected large memory cost. In addition, it is also difficult to achieve the balance between memory cost and distributed training accuracy because more duplicated information will bring higher training accuracy but incur low training speed. 
Therefore, in this work, we will solve the research challenge to achieve the optimal balance of higher accuracy and faster training speed.

\section{GAD Framework }

In this section, we describe training procedure of graph augmentation-based distributed GCN framework (GAD). We first introduce the overall framework of GAD in Section \ref{method:1}, followed by GAD-Partition in Section \ref{method:2}, and finally present the distributed GCN training and its optimization in Section \ref{met:distributed training} and Section \ref{method:var measurement} respectively.

\subsection{Framework overview}\label{method:1}
\begin{figure}
\centering
\includegraphics[scale=0.58]{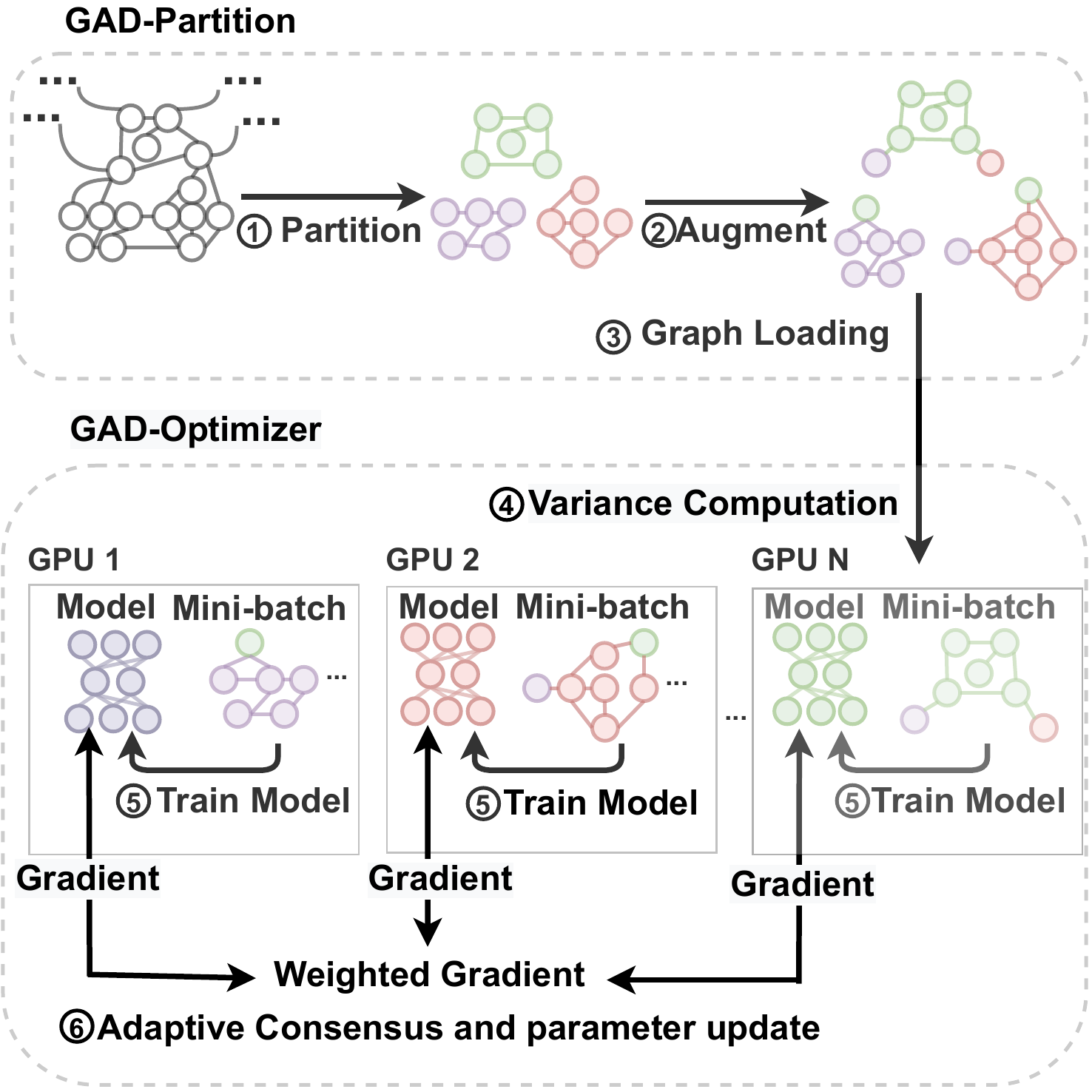}
\caption{Framework Overview}
\label{fig:Model Overview}
\end{figure}



As illustrated in Figure \ref{fig:Model Overview}, GAD framework is divided into two key components: GAD-Partition and GAD-Optimizer. The working procedure of both GAD-Partition and GAD-Optimizer is described in six steps.  (i) To divide a graph $\mathcal{G}$ into subgraphs $g_i\in\{g_1 \dots g_k\}$, we extend the robust graph partition method Metis\cite{karypis1998fast} that minimizes the number of edges across subgraphs, i.e., candidate replication nodes can be minimize for each $g_i$. 
(ii) Propose a novel subgraph augmentation approach to select and replicate the important nodes across subgraphs from candidate replication nodes to graph $g_i$. The augmented subgraph $g_i'$ guarantees high training accuracy with least number of replicated nodes.
(iii) Simple graph loading is to assign the augmented subgraphs to different processors $\{ p_1 \dots p_n \}$. 
(iv) Graph variance $\operatorname{Var}(g_i')$ for $g_i'$ is a significant factor to engage the convergence of training propagation. Unlike, traditional graph variance $\operatorname{Var}(g_i')$ that generates subgraphs by sampling, we present graph variance-based importance $\zeta$ to assess the degree distribution of $g'$ and the difference of node embedding in $g'$ generated by partition. (v) Distributed GCN training is performed by calling single GCN model at $p_i$ in $\{ p_1 \dots p_n \}$ and (vi) present weighted global consensus mechanism to aggregate the gradients of GCN model with the graph variance-based importance $\zeta$.

\subsection{GAD-Partition}\label{method:2}

The GAD-Partition aims to divide the graph into several roughly equal partitions and extend the partitions by replicating the nodes with significant communication. A straightforward solution is to replicate the communication nodes evenly; however, it requires manually adjust the number of replication nodes to achieve low redundancy and high accuracy, i.e., higher replication nodes number does not impact the accuracy but performance due to redundancy.

 
 To overcome this issue and achieve high accuracy under minimum redundancy, we first employ graph partitioning that minimizes the candidate replication nodes. Then, we propose a Monte-Carlo-based node importance measurement and depth-first sampling method to replicate high-quality nodes with structural regularities from the candidate replication nodes. Finally, we assign the subgraphs to different processors. Following, we describes the procedure in detail.

\begin{figure}[h]
\centering
\includegraphics[scale=0.58]{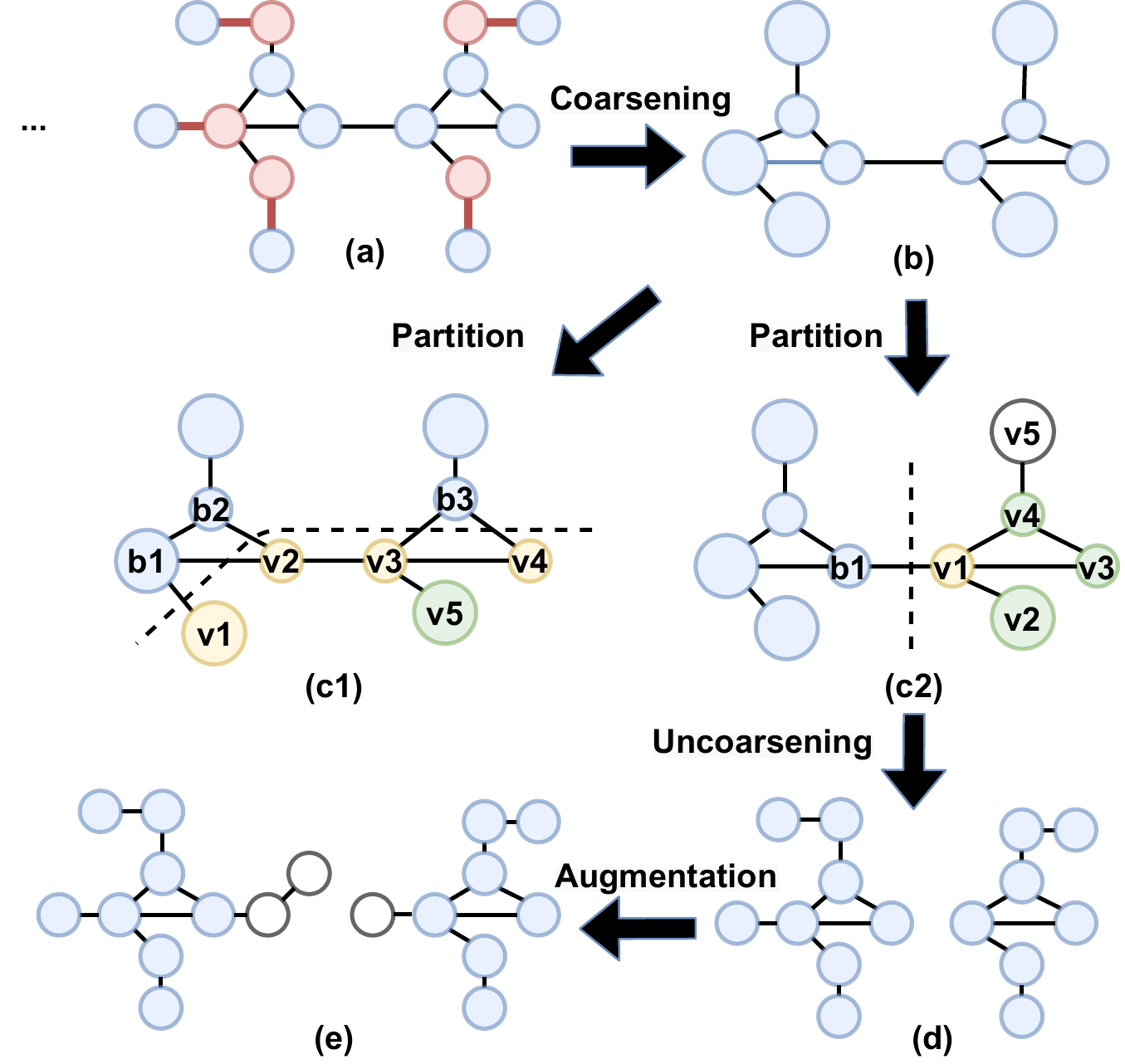}
\caption{Overview of GAD-Partition}
\label{fig:k-hop_replicate}
\end{figure}

\subsubsection{Graph Partition}
Graph partitioning divides a graph into several disjoint sub-graphs. It can be defined as follows.

\begin{definition}[{Graph partition}]\label{def:Multi-level graph partition}
Given an undirected graph $\mathcal{G} = (V, E)$, graph partition divides $\mathcal{G}$ into $k$ disjoint subgraphs $g_1 \dots g_k$ with nodes sets $V_1,\dots ,V_k$ and edges sets $E_1, \dots E_k$ such that $\bigcup_{i=1}^k V_i = V$. The size of $V_i$ and $E_i$ is denoted by $|V_i|$ and $|E_i|$ respectively.
\end{definition}

To minimize the number of candidate replication node in each subgraph and keep the balance of computational load in the distributed environment, our target is to divide a graph satisfying:

\begin{equation}\label{partition edges limitation}
    min \left(|E| - \sum_{i=1}^{k}|E_i|\right)
\end{equation}

subject to

\begin{equation}\label{partition nodes limitation}
    |V_i| \leq (1+\epsilon) \left \lceil \frac{|V|}{k} \right \rceil
\end{equation}


Where $k$ is the number of partitions, $\epsilon$ is the imbalance constant, indicating the tolerable difference of node numbers over the partitions. Intuitively, Eq. \ref{partition edges limitation} 
minimizes edge cuts between subgraphs, and Eq. \ref{partition nodes limitation} ensures the approximate balance for the number of nodes on each subgraph. Here, candidate replication node can be defined as follows:

\begin{definition} [{Candidate replication node}]\label{def:Candidate replication node}
Given a subgraph $g_i = (V_i, E_i)$ and an original graph $\mathcal{G} = (V, E)$, the candidate replication nodes of $g_i$, denoted as $C(g_i)$, are the $x$-hop neighborhood for each node in boundary nodes $B(g_i) \in \left\{v \mid v \in V_i \wedge \exists e_{v, u} \in E \backslash E_i\right\}$, where $C(g_i) \notin V_i$ and $x$ is equal to the number of GCN layer.
\end{definition}

To balance the number of nodes across the partitioned subgraphs (Eq.\ref{partition nodes limitation}) and minimize the number of candidate replication nodes (Eq. \ref{partition edges limitation}), inspired by Metis\cite{karypis1998fast}, we use the multi-level graph partitioning method. Below, we describe the implementation procedure for proposed graph partitioning.

\begin{enumerate}
 
\item Coarsening Phase: 
To reduce the graph size while maintaining its important properties, we iteratively contracts the graph $\mathcal{G} = (V, E)$ to coarsened graph $\mathcal{G}_c = (V_c', E_c')$ such that $\mathcal{G}_c \subseteq \mathcal{G}$. For graph $\mathcal{G}$, we initialize every node's and edge's weight to 1. 
 To generate the coarsened graph $\mathcal{G}_c$, we randomly select the number of nodes $\{\hat{v}_1 \dots \hat{v}_m\}$ from $\mathcal{G}_{c-1}$. For each $\hat{v}_i$ in $\{\hat{v}_1 \dots \hat{v}_m\}$, we find its adjacent edge with the maximum weight, e.g., the edge is denoted as $(\hat{v}_i, \hat{v_i}')$. If there are multiple adjacent edges that have the same maximum weight, we randomly select one of them. Then, we merge $\hat{v}_i$ and $\hat{v_i}'$ as a coarsened node where the weight of the coarsened node is the aggregation of two merging nodes' weight, and its connected edge's weight is the sum of the edges to be merged. We repeat the above steps to continuously coarsen the graph until the number of nodes in $\mathcal{G}_c$ is reduced to a certain value, e.g., 20\% number of nodes in the original graph. 

\item Partition Phase: For a given coarsened graph $\mathcal{G}_c$, we divide $\mathcal{G}_c$ into $k$ coarsened partitions $\hat{g_1} \dots \hat{g_k}$ based on the constraints described in Eq. \ref{partition edges limitation} and Eq. \ref{partition nodes limitation}. To do so, firstly we randomly choose $k$ nodes from the current level $\mathcal{G}_c$ as the seeds of k partitions $\hat{g_1} \dots \hat{g_k}$. Then, for each partition $\hat{g_i} \in \{\hat{g_1} \dots \hat{g_k}\}$, we expand the partition $\hat{g_i}$ by adding the nodes that satisfy (i) these nodes are not contained in $\hat{g_i}$ and have the edges linked to a node of $\hat{g_i}$; (ii) the edges between the added nodes and the nodes of $\hat{g_i}$ have the maximum weight. The above procedure is repeated until the sum of node weights in each partition satisfies the constraint in Eq. \ref{partition nodes limitation}. 
After that, there may have some nodes that are not belong to any partitions. To solve this problem, we pick up each node and add it into the nearest partition. In order to achieve the goal in Eq. \ref{partition edges limitation}, we run the above procedure for many times and obtain multiple partition candidate results. Finally, we select the approximated optimal result by taking the result with the minimum $|E| - \sum_{i=1}^{k}|E_i|$, denoted as the coarsened partitions $\hat{g_1} \dots \hat{g_k}$.



\item Uncoarsening Phase: For each coarsened subgraph $\hat{g_i} \in$ \{$\hat{g_1}, \dots , \hat{g_k}$\}, we obtain its corresponding partition $g_i$ by uncoarsening the coarsened nodes in $\hat{g_i}$. In other words, we can get the graph partitions \{$g_1$, ..., $g_k$\} of the original graph.


\end{enumerate}

\begin{example}[{Multi-level graph partition}]

Figure \ref{fig:k-hop_replicate}.a - Figure \ref{fig:k-hop_replicate}.d show how to minimize the number of candidate replication nodes across the different subgraphs. Coarsening process is illustrated in Figure \ref{fig:k-hop_replicate}.a - Figure \ref{fig:k-hop_replicate}.b. Randomly selected nodes and their maximum weighted adjacent edges are highlighted by red color in Figure \ref{fig:k-hop_replicate}.a.  Figure \ref{fig:k-hop_replicate}.b - Figure \ref{fig:k-hop_replicate}.c1 indicate the coarsening-node with larger size. Figure \ref{fig:k-hop_replicate}.b - Figure \ref{fig:k-hop_replicate}.c1, Figure \ref{fig:k-hop_replicate}.b - Figure \ref{fig:k-hop_replicate}.c2 indicate two partition results with 5 and 1 edges cuts, respectively. Compared to Figure \ref{fig:k-hop_replicate}.b - Figure \ref{fig:k-hop_replicate}.c2, we can see the result of Figure \ref{fig:k-hop_replicate}.b - Figure \ref{fig:k-hop_replicate}.c1 with more edge cuts has more 1-hop (marked as yellow) and 2-hop (marked as green) candidate replication nodes. Thus, we choose the partition result with less edge cut to minimize the number of candidate replication nodes. Figure \ref{fig:k-hop_replicate}.c2 - Figure \ref{fig:k-hop_replicate}.d show the process of uncoarsening the coarsened graph with partitions.

\end{example}

\subsubsection{Local Subgraph Augmentation}


After obtaining subgraphs, we propose a strategy to add high-quality replication nodes with structural regularities for each subgraph according to the importance measurement of each candidate replication node and the depth-first selection from candidate replication nodes.


\noindent\underline{\textbf{Importance measurement:}} To assign an importance weight to each candidate replication node, the common practice considers the structure around the node, i.e., use the degree of the node as importance weight. However, it does not work in our case, as candidate replication nodes are also associated with different parts of the subgraph. We use the random walk to capture the structure information and dependencies between candidate replication nodes and subgraphs to overcome the challenge mentioned above. In addition, we use the Monte-Carlo approach to approximate the occurrence frequency of each candidate replication node in random walks. In specific, for each subgraph $g_i=(v_i, e_i)$, we randomly select a node from boundary node-set $B(g_i)$ of $g_i$ as the beginning of random walk (RW) and form a RW sequence $RW_j$. We repeat this operation $n$ times and traverse RW sequences to calculate important weight $I(v)$ for each $v \in C(g_i)$, which can be measured as follows:

\begin{equation} \label{for:pro}
I(v) = \frac{\sum_{j=1}^n RW_j(v)}{n}, \quad v \notin v_i
\end{equation} 

Where $RW_j(v)=1$ if $v$ in the $j$-th RW sequence, otherwise $RW_j(v)=0$. 

To identify the length $l$ of RW sequence, we present a property as below.

\begin{property}\label{pro:step}
Given a subgraph $g_i=(v_i, e_i)$, all candidate replication nodes of $g_i$ are covered by RW sequences without irrelevant nodes when the length of RW sequence $l$ is equal to the number of GCN layer. 
\end{property}
\begin{IEEEproof}
If we set RW with length $l$, it can reach to all the $l$ hop neighbors. Based on Definition \ref{def:Candidate replication node}, if $l$ is equal to the number of GCN layer and starts from the nodes in $B(g_i)$, RW can reach all the candidate replication nodes. Thus, $l$ is equal to the number of GCN layer.


\end{IEEEproof}

To converge $l(v)$ to acceptable range in minimum time, we need to minimize $n$ in Eq. \ref{for:pro}. In essence, Eq. \ref{for:pro} is a kind of Monte-Carlo method, which simulates the probability distribution of sampling through multiple enumerations. Thus, we can estimate the enumeration times based on the Monte-Carlo error Eq. \cite{Driels2004DeterminingTN}. Formally, Monte-Carlo error $E$ can be measured as follows:

\begin{equation}\label{mk_error}
\mathrm{E}=\frac{z_{c} \sigma_{x}}{\bar{x} \sqrt{n}}
\end{equation} \
Where $z$ is a statistic number, representing a certain confidence interval.  We set the 95\% confidence level and $E=0.05$ to ensure that the estimation is close to the true value. It can be described as, for 95\% confidence interval, the estimation will not differ by more than 5\% from the true. When we have 95\% confidence level, we can get the value of $z_c$ from Z-test that $z_c=1.96$. $\sigma_{x}$ and $\bar{x}$ are the standard deviation and the mean for the sample, respectively.

The detailed procedure is presented in Algorithm \ref{approximation} (Line 1-17). We iteratively choose a subgraph $g_i$ from partitions (Line 1). For each $g_i$, the RW sequence set $R$ is initialized as empty (Line 2). Then we add a small number of RW sequences to $R$ (Line 3-8), and estimate the mean $\bar{x}$ and variance $\theta$ of $I(v)$ for $v$ in $R$ (Line 9-10). Next, we calculate the value of repeat time $n$ by using Eq. \ref{mk_error} with $\bar{x}$ and $\sigma$ (Line 11). Finally, we add the rest number of RW sequences to $R$ (Line 12-16) and calculate $I(v)$ for each candidate replication nodes (Line 17).

\noindent\underline{\textbf{Importance based augmentation:}} To overcome the aforementioned challenge of node replication (balance high accuracy and low redundancy), we introduce graph density $d(g_i)$ to compute the number of replication nodes $n(g_i)$ for subgraph $g_i$. 


\begin{definition} [{Graph density}]\label{def:subgraph density}
For a given subgraph $g_i$ with the nodes number $|v_i|$ and edge number $|e_i|$, the density of $g_i$ can be computed as:

\begin{equation}
d(g_i) = \frac{2|e_i|}{|v_i|(|v_i|-1)}
\end{equation}
\end{definition}

\begin{figure}
\centering
\includegraphics[scale=0.7]{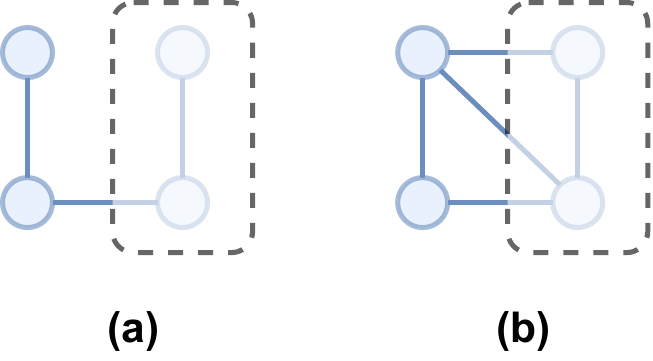}
\caption{Relationship between density and information loss}
\label{fig:density}
\end{figure}

\begin{example}
 As shown in Figure \ref{fig:density}, high density subgraph (Figure \ref{fig:density}.b with 3 edges cut) has higher information loss than low density subgraph (Figure \ref{fig:density}.a with 1 edge cut) under same partition. Therefore, more nodes need to be replicated for density graph to maintain the accuracy.
\end{example}

Notice that graph density $d(g_i)$ ranges from 0 to 1. In general, it can be treated as the complement of replicated node. Here, we determine the number of replicate nodes $n(g_i)$ according to $d(g_i)$, which can be measured as below.

\begin{equation}
n(g_i) = \alpha\left(1 + d(g_i)\right)|v|
\end{equation}
where $\alpha$ is the constant, i.e., $\alpha=0.01$.

We need to consider how to select important nodes on the basis of $I(v)$. The naive solution is to copy top $n(g_i)$ most important replication nodes. But it may have dangling nodes, which means that the selected replication nodes have no path to $g$. To solve this problem, we propose a depth-first selection strategy. The detailed procedure is presented in Algorithm 1 (Line 18-26). Firstly, we calculate the $n(g_i)$ for $g_i$ (Line 18), then compute the important score $I(RW_i)$ for each $RW_i$ through sum $I(v)$ for $v$ in $RW_i$, and iteratively select $RW$ with maximum $I(RW)$ (Line 20). Next, we add nodes $v$ in $RW$ to the replication node set $v'$ that never appear in $v'$ until the number of replication node is equal to $n(g_i)$ (Line 21-25). Finally, we add $v'$ back to subgraph $g'$ with the connected edges $e'$ (Line 26).

\begin{algorithm}[h!]
	\caption{Importance-based subgraph augmentation}
	\begin{algorithmic}[1] 
 		\Require Subgraphs $ {g}_1, {g}_2, \dots , {g}_k$, Original graph $\mathcal{G}$, Number of GCN layer $l$
		\Ensure Augmented subgraphs $ {g_1}', {g_2}', \dots , {g_k}'$
        \For{$i \gets 1$ to $k$}
            \State $R \gets \emptyset$
            \State $d$=ave\_degree($B(g_i)$)
            \For{$j \gets 1$ to $d|B(g_i)|$}
                \State $v_j$ = Randomly select boundary node from $g_i$
                \State $O=v_j$.random\_walk(step=$l$,graph=$G$)
                \State $R$.push($O$)
            \EndFor
            \State $prob$ = $R$.bincount() / $R$.sum()
            \State $\sigma, \bar{x}$ = $prob$.var(), $prob$.mean()
            \State $n = (z_{c} \sigma_{x} / \bar{x} \mathrm{E})^2$
            \For{$j \gets d|B(g_i)|$ to $n$}
                \State $v_j$ = Randomly select boundary node from $g_i$
                \State $O=v_j$.random\_walk(step=$l$, graph=$G$)
                \State $R$.push($O$)
            \EndFor
            \State $I$=argsort(sum($R$.bincount()))
            \State $n(g_i) = \alpha \left(1 + d(g_i)\right) |v_i|$
            \Repeat 
                \State $RW$ = $R$.pop(index=I.pop())
                \State $v_i' \gets \emptyset$, $e_i' \gets \emptyset$
                \If {$v$ in $RW \notin v'$}
                    \State $v_i'$.push($v$),\quad$e_i'$.push($e_{i,g}$)
                \EndIf
            \Until{$|v'| = n(g_i)$}
            \State ${g_i}' = (v_i + {v_i}', e_i + {e_i}')$
        \EndFor
        \State \Return $ {g_1}', {g_2}', \dots , {g_k}'$;
	\end{algorithmic}
 \label{approximation}
\end{algorithm}

\subsubsection{Subgraph Loading}\label{sec:subgraph loading}
To allocate augmented subgraphs to the multiple processors and achieve the workload balance in the distributed GCN training, we allocate a similar number of nodes to each processor. For the given augmented subgraph sequence ${g_1', \dots , g_k'}$ and processors ${p_1, \dots, p_n}$, we iteratively select and allocate $g_i$ to $p_i$ if $p_i$  has the least number of nodes.

\subsection{Distributed GCN training}\label{met:distributed training}

In this section, we describe the process of standard distributed GCN training. Following \cite{DBLP:journals/tsipn/ScardapaneSL21}, we first do forward propagation in each processor, which is measured as follows:

\begin{equation}\label{def:forward1}
H^{(l+1)}_{p} = \sigma\left(\widetilde{A}{\cdot}H^{(l)}_{p}{\cdot}W^{(l)}_{p}\right)
\end{equation}
where $l$ stands for GCN layer, $\sigma$ is activation function, such as ReLU or LeakyReLU, $p$ is the $p$-th processor, $W^{(l)}$ is the trainable weight matrix of $l$. Prediction $\hat{y}$ can be calculated by the forward propagation stacking. Specifically, a two layer stacking for calculating $\hat{y}$ can be denoted as below.
\begin{equation} \label{def:forward2}
\hat{y} = softmax\left(\tilde{A}{\cdot}\sigma\left(\tilde{A}{\cdot}H^{(0)}{\cdot}W^{(1)}\right){\cdot}W^{(2)}\right).
\end{equation}

Secondly, we execute the backward propagation to calculate the loss and gradient at each processor. The first loss layer can be measured as follows:
\begin{equation} \label{def:backward1}
\mathcal{L}=-\sum_{i=1}^{N} y^{(i)} \log \hat{y}^{(i)}+\left(1-y^{(i)}\right) \log \left(1-\hat{y}^{(i)}\right)
\end{equation} 
For gradient, it can be denoted as 
\begin{equation}\label{def:gradient}
\triangledown W = \frac{\partial \mathcal{L}}{\partial \hat{y}}=(-\frac{y_{x}}{\hat{y}}+\frac{y_{j}}{1-\sum_{i=1, t j}^{k} \hat{y}}) 
\end{equation} where $\frac{\partial \mathcal{L}}{\partial \hat{y}}$ stands for the gradient $\hat{y}$ under $\mathcal{L}$. After obtaining the gradient in each processor, the distributed GCN training will aggregate all the gradients and send the average gradient back to each processor for parameter update. This process is called \emph{global consensus}. 

\begin{definition}[{Global consensus}\cite{DBLP:journals/tsipn/ScardapaneSL21}] \label{def:global consensus}

Given a gradient sequence $\{ \triangledown W_1, \triangledown W_2, \dots \triangledown W_n \}$ calculated by GCN models in processors $\{ {p_1},$ ${p_2}, \dots ,{p_n} \}$, the global consensus can be measured as follows:

\begin{equation} \label{eq:global consensus} \triangledown W = \frac{1}{ n } (\triangledown W_1 + \triangledown W_2 + \dots, \triangledown W_n) \end{equation}

\end{definition} 

Finally, we use $\triangledown W$ and learning rate $\eta$ to update the parameters of GCN models in each $p_i$ synchronously, which can be denoted as:

\begin{equation}\label{method:update}
W_{i}={W_i} - \eta \frac{\partial \mathcal{L}}{\partial \triangledown W}
\end{equation}

After updating the parameters, the distributed GCN training framework synchronously selects the next subgraph as a mini-batch to train GCN models in each processor until all the subgraphs are traversed. We repeat this process until the loss function converges.

\subsection{GAD-Optimizer}\label{method:var measurement} 


For any training method of GCN, we should consider the problem of slow convergence and poor generalization caused by the variance of subgraphs. A general approach is to reduce the subgraphs variance by optimizing the sampling node selection strategy \cite{DBLP:conf/iclr/ChenMX18, DBLP:conf/nips/Huang0RH18, DBLP:conf/iclr/ZengZSKP20}, but it can not be applied to GAD-Partition because graph partition has different ways to form subgraphs. To solve this problem, we propose the GAD-Optimizer, which reduces the impact of high variance subgraphs by assigning a low importance weight to the trained gradient of the subgraph. Here we separate GAD-Optimizer in two parts: variance-based subgraphs importance and weighted global consensus.


\subsubsection{Variance-based Subgraph Importance}

To measure the subgraph variance $\operatorname{var(g)}$, we follow the approach described in \cite{DBLP:conf/iclr/ZengZSKP20} for sampling-based subgraph $g$, which can be measured as follows:


\begin{equation}\label{equ:var}
\operatorname{var}(g) = \sum_{\ell} \sum_{i, j \in V} \frac{H_{j}^{(\ell)}}{p_{i}\alpha_{j, i}} d(i,j)
\end{equation}

Where $l$ is the $l^{th}$-layer in a GCN model, $p_i$ is the probability to select the node $v_i$, $\alpha_{u, v}$ is the normalization constant and calculated as $\alpha_{u, v}$=$\frac{C_{u, v}}{C_{v}}$. $C$ is the node sampling times, $d(i, j)$ is the Euclidean distance between nodes $v_i$ and $v_j$.

Intuitively, the variance of the sampling-based subgraph is calculated by accumulating the variance between two nodes in the sampling layers. Since the subgraphs in our work are generated by partition instead of sampling, we can get $\operatorname{var}(g)$ directly by accumulating the variance between two nodes in $g$. Thus, to compute  $\frac{H_{j}^{(\ell)}}{p_{i}\alpha_{j, i}}$ in Eq. \ref{equ:var}, we only need to consider the difference of node degree without the dependence of nodes during sampling. We provide the following property to measure the difference of node degree for subgraphs generated by the partition method.

\begin{property}\label{property:2} Given a subgraph $g_i$ and its node selection probability $p(v)$, $\sum_{i, j\in v}p(v_i)p(v_j)$ is large if the difference of node degree in $g_i$ is small.
\end{property}

\begin{IEEEproof}
Consider we have $n$ nodes such that $V=\{v_1,\dots,v_n\}$. Assume we have the probability $p(v_1)=\frac{1+t}{n}$, $p(v_2)=\frac{1+t}{n}$, $\dots$, $p(v_{\frac{1}{2}n})=\frac{1-t}{n}$, $\dots$, $p(v_n)=\frac{1-t}{n}$ for each node so that $\sum_{v_i \in V}p(v_i)=1$ in a special case. 
To maximize $\sum_{i, j\in v}p(v_i)p(v_j)= p(v_1)p(v_2)+p(v_1)p(v_3)+\dots+p(v_{n-1})p(v_{n})$, requires to maximize $\left( \left(\frac{n}{2}-1\right) + \left(\frac{n}{2}-2\right) + \dots + 1\right)\left([\frac{1+t}{n}]\cdot[\frac{1+t}{n}]\right) + \frac{n^2}{4} \left([\frac{1+t}{n}]\cdot[\frac{1-t}{n}] \right) + \left( \left(\frac{n}{2}-1\right) + \left(\frac{n}{2}-2\right) + \dots + 1\right)\left([\frac{1-t}{n}]\cdot[\frac{1-t}{n}]\right)$
, which can be simplified as $i\frac{1}{n^2}$ $-$ $\frac{n}{2}t^2$. Since $n$, $i$ are constants, thus, to maximize $\sum_{i, j\in v}p(v_i)p(v_j)$, it is required to minimize $\frac{n}{2}t^2$, which can be achieved when $t=0$. So the $\sum_{i, j\in v}p(v_i)p(v_j)$ becomes larger when we have small difference of $p(v_i)$.
\end{IEEEproof}

According to Property \ref{property:2}, we replace $\frac{H_{j}^{(\ell)}}{p_{i}\alpha_{j, i}}$ to $\sum_{i, j\in v}p(v_i)p(v_j)$ in Eq. \ref{equ:var}. In addition, since the difference of node degrees is inversely proportional to the variance-based important weight $\zeta$, we change it from denominator to numerator. To this end, we can write $\zeta$ as:

\begin{equation}\label{equ:important weight}
\zeta(g') = \frac{\sum_{i, j\in v}p(v_i)p(v_j)}{d(i,j)+\beta}
\end{equation}

Where $\beta$ is the constant to limit the denominator from being zero. Normally $\beta = 1$. $\zeta({g}')$ can be calculated in distributed environment during subgraph loading process (described in section \ref{sec:subgraph loading}). In addition, $v_i, v_j$ in Eq. \ref{equ:important weight} can be extended to multiple dimensions feature by calculating each dimension separately.  To measure $\zeta({g}')$ of subgraph $g'$, following, we illustrated with an example.

\begin{figure}[h!]
\centering
\includegraphics[scale=0.70]{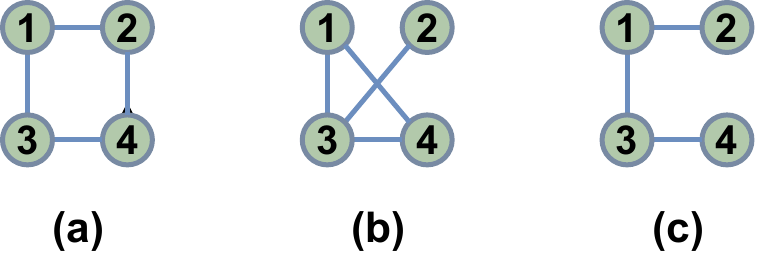}
\caption{Example of variance-based important weight.}
\label{fig:var}
\end{figure}

\begin{example}


Figure \ref{fig:var} demonstrates the variance-based important weight of three subgraphs 
with different degree distribution. Notice that, the distance of the nodes in each subgraph is zero, i.e., $d(i,j)=0$ for any two nodes $v_i$, $v_j$ in a subgraph. We can observe that Figure 4.b has the highest variance with degree sequence (3, 2, 2, 1). Whereas, the subgraph in Figure 4.a has the lowest variance with sequence  (2, 2, 2, 2). According to Eq. (\ref{equ:important weight}), $\zeta(g')$ holds  $3.75>3.61>3.59$ for Figure 4.a, Figure 4.c, Figure 4.b,  respectively.

\end{example}

\subsubsection{Weighted Global Consensus}
To reduce the impact of variance for the subgraphs generated by GAD-Partition, we extend the global consensus (Definition \ref{def:global consensus}) by providing less weight to subgraphs with higher variance. Specifically, Eq. \ref{eq:global consensus} can be rewritten as:

\begin{equation}\label{method:average weight} \triangledown \hat{W} = \frac{1}{ \sum_{j=1}^{n} \zeta_j } \sum_{i = 1}^{n} ({\zeta_1}\triangledown W_1 + {\zeta_2}\triangledown W_2 + \dots, {\zeta_n}\triangledown W_n) \end{equation}


To update parameters of GCN, we can rewrite  Eq. \ref{method:update} as: 

\begin{equation}\label{eq:new update}
W_{i}={W_i} - \eta \frac{\partial \mathcal{L}}{\partial \triangledown \hat{W}}
\end{equation}


We provide the distributed GCN training and GAD-Optimizer process in Algorithm \ref{consensus}. Line 1-4 indicates the process of subgraph allocation and initialization. Line 5-9 describes distributed GCN training. Finally, Line 10-12 describes the weighted global consensus and parameter update.

\begin{algorithm}[htpb]
	\caption{Distributed GCN training by weighted global consensus}
	\begin{algorithmic}[1] 
		\Require Augmented graphs $ {g_1}', {g_2}', \dots , {g_k}'$ where ${g_i}'=(v_i', e_i')$, Labels $\boldsymbol{\bar{y}}_v$, Processors $\mathcal{P}=\{p_1, p_2, \dots , p_n\}$, GCN models for each processor $\mathcal{M} = \{m_1, m_2, \dots, m_n\}$
		\Ensure Updated models $\mathcal{M'} = \{m_1', m_2', \dots, m_n'\}$
		\State $\tilde{v}$=average($v_1'+v_2' + \dots + v_k'$),       $\tilde{e}$=average($e_1'+e_2' + \dots + e_k'$)
            \For{$i \gets 1$ to $n$}
                \State Assign $z$ graph $\{g_j, \dots, g_k\}$ to $p_i$ which $|v_j+ \dots +v_k| \sim z|\tilde{v}|$, $|e_j+ \dots +e_k| \sim z|\tilde{e}|$
            \EndFor
        
        \For{each processor $p$ $\mathbf{distributed}$ $\mathbf{parallel}$}
            \For{Each batch $g_k$ in $p_i$}
                \State $\{\boldsymbol{y}_v | v \in v_k\} \gets$ Forward propagation $\{\boldsymbol{x}_v | v \in v_k\}$ using Eq. \ref{def:forward2}
                \State $\triangledown W_i \gets$ Backward propagation from loss $L(\boldsymbol{y}_v, \boldsymbol{\bar{y}}_v)$ using Eq. \ref{def:backward1}, \ref{def:gradient}
            \EndFor
            \State Calculate $\triangledown \hat{W}$ using Eq. \ref{method:average weight}
            \State $m_i'$ = Update parameter of $m_i$ using Eq. \ref{eq:new update}
        \EndFor
        \State \Return $\{m_1, m_2, \dots, m_n\}$;
	\end{algorithmic}
 \label{consensus}
\end{algorithm}



\section{Experiments}
This section first presents experimental setups such as datasets, baseline methods, and parameter configuration, followed by an analysis the GAD framework's effectiveness, accuracy, and stability and the effect of graph augmentation and weighted global consensus. To eliminate the impact of different parallelization architectures on training time, we implemented our framework and all baseline models under the single thread, single process, and multiple GPU processors.  
Experiments are implemented using Pytorch 1.5.0 with Python 3.7.7 and ran on four NVIDIA GeForce GTX 1080 Ti 11GB Memory with no NVLink connections. 





\subsection{Experiment Setup}\label{exp:setup}
\noindent\textbf{\underline{Datasets:}} We performed extensive experiments on four real-world benchmark datasets to demonstrate the effectiveness of our distributed GCN optimization framework. Table \ref{table:dataset} shows the statistics for all the datasets.

\begin{table*}[htpb]
    \caption{Dataset Statistics}
    \vspace{5pt}
    \centering
    \begin{tabular}{c c c c c c}
        \hline
        Dataset &   Nodes   &   Edges   &   Labels  &   Features    &   Data splits of Training/Validation/Test (\%) \\
        \hline
        Cora  &   2,708   & 5,429   &   7  &   1433 &   45/18/37 (\%)  \\
        Pubmed  &   19,717   & 44,324   &   3  &   500 &   92/03/05 (\%)  \\
        Flicker  &   89,250   & 899,756   &   7  &   500 &   50/25/25 (\%)  \\
        Reddit  &   231,443   & 11,606,919   &   41  &   602 &   70/20/10 (\%)  \\
        \hline       
    \end{tabular}
    \label{table:dataset}
\end{table*}

\footnotetext[1]{\url{https://pytorch-geometric.readthedocs.io/en/latest/modules/datasets.html}}

\begin{enumerate}

\item[$\bullet$] \textbf{Cora}\footnotemark[1] is a citation network with the scientific papers as the nodes, and their citation relationships as the edges. Here, each node is represented by a 1433-dimensional word vector. 

\item[$\bullet$] \textbf{Pubmed}\footnotemark[1] is a citation network. The nodes are composed of scientific publications on diabetes, represented by tf-idf word vector in a dictionary of unique words. The edges represent the citation relationships of the publications.

\item[$\bullet$] \textbf{Flicker}\footnotemark[1] is an image network. The nodes are composed of online images, represented by a 500-dimensional descriptions vector. The edges represent the relationship between images.

\item[$\bullet$] \textbf{Reddit}\footnotemark[1] is a user interaction network. The nodes are composed of posts on the Reddit forum, represented by the time and title vector of the user comment. The edges indicate that the same person comments on two posts.


\end{enumerate}

\noindent\textbf{\underline{Baseline Methods:}} We compared our framework with the following baseline methods implemented in a distributed environment.
\begin{enumerate}
    \item[$\bullet$] 
\textbf{Distributed GCN}\cite{DBLP:conf/iclr/KipfW17} is a classical graph network model. It uses a convolution computation to gather information of neighbour nodes randomly within a graph.

  \item[$\bullet$] \textbf{Distributed GraphSAGE}\cite{DBLP:conf/nips/HamiltonYL17} is a framework for inductive representation learning on large graphs. It gathers neighbor information by sampling a fixed number of neighbor nodes uniformly and randomly and provides a strategy to learn the different aggregation functions on a different number of hops.
  
  
    \item[$\bullet$] \textbf{Distributed Cluster-GCN}\cite{DBLP:conf/kdd/ChiangLSLBH19} is a training method for scalable training of graph network model using stochastic gradient descent. It samples a dense subgraph identified by a graph clustering algorithm, and restricts the neighborhood search within this subgraph.
    
  
        \item[$\bullet$] \textbf{Distributed GraphSAINT}\cite{DBLP:conf/iclr/ZengZSKP20} is a graph sampling-based inductive learning method. It constructs mini-batches by sampling the training graph. Here, we compare its three sampling strategies and indicate them as GraphSAINT-Node, GraphSAINT-Edge, and GraphSAINT-RW. Specifically, GraphSAINT-Node is the layer-based random node sampling, GraphSAINT-Edge is the layer-based random edge sampling, and GraphSAINT-RW is the random walk sampling.
\end{enumerate}

\noindent{\textbf{\underline{Parameter Configuration:}}}\label{Parameter Configuration}
We have considered four parameters in our experiments: learning rate $\eta$, GCN layer number $l$, batch size $b$, and number of hidden neurons $h$. In order to compare the performance with baselines, we set the default learning rate $\eta = 0.0001$ for Cora, Flicker, Reddit datasets, and $\eta = 0.001$ for Pubmed dataset. In addition, we set the default batch size $b=300$ for Cora, Flicker, Reddit datasets, and $b=1500$ for Pubmed dataset. To determine the GCN layer number $l$ and hidden number of neurons $h$, we use the optimal parameter settings for each experiment by varying $l$ and $h$ such that $ l \in \{2,3,4\}$ and $h \in \{128, 256, 512\}$.

\subsection{Evaluation of Effectiveness and Accuracy}\label{exp:Evaluation of Effectiveness and Accuracy}

\begin{table*}[htbp]
    \caption{Compared accuracy of different GCN training methods}
    \vspace{5pt}
    \centering
    \begin{tabular}{ll c c c c c}
        \hline
        Method &  Cora  &  Pubmed   &   Flicker   &   Reddit  \\ 
        \hline
        Distributed GCN     & 0.5525  &  0.7117  & 0.4769     &   0.7433  \\
        Distributed GraphSAGE   & 0.5727 & 0.7199  &   0.4689   &  0.7653   \\
        Distributed ClusterGCN  & 0.8093 & 0.7511  &  0.4886  &   0.9272    \\
        Distributed GraphSAINT-Node  & 0.7450  & 0.6585  &   0.4534   &  0.8921  \\
        Distributed GraphSAINT-Edge  & 0.4027 & 0.7459  &   -   &  -  \\
        Distributed GraphSAINT-RW  & 0.7463  & 0.7445  &   0.4546   &  0.9152  \\
        \hline
        GAD  & \textbf{0.8160}  &   \textbf{0.7558}   & \textbf{0.4907}  &  \textbf{0.9313}  \\ 
        \hline   
    \end{tabular}
    \label{tb:exp best model accuracy}
\end{table*}

\begin{table*}[htbp]
    \caption{Stability of GAD on test accuracy when GPU and layer number vary.}
    \vspace{5pt}
    \centering
    \begin{tabular}{c c c c c c}
        \hline
        GPU Number &   2 Layers   &   3 Layers   &   4 Layers  \\
        \hline
        \textbf{1 GPU}    &   \textbf{0.7443 }   &   \textbf{0.7398 }   & \textbf{0.7540}  \\ 
        2 GPUs    &   0.7378    &   0.7330   &  0.7518  \\
        3 GPUs    &   0.7428   &    0.7304   &  0.7490  \\ 
        4 GPUs    &   0.7369    &   0.7345   &  0.7503 \\
        \hline   
    \end{tabular}
    \label{tb:layers and GPU numbers accuracy comparision}
\end{table*}

\begin{figure}[htbp] 
\centering
\includegraphics[scale=0.44]{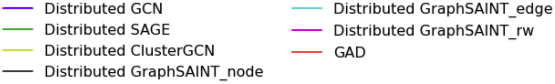} \hfill
\subfloat[Cora]{{\includegraphics[scale=0.39]{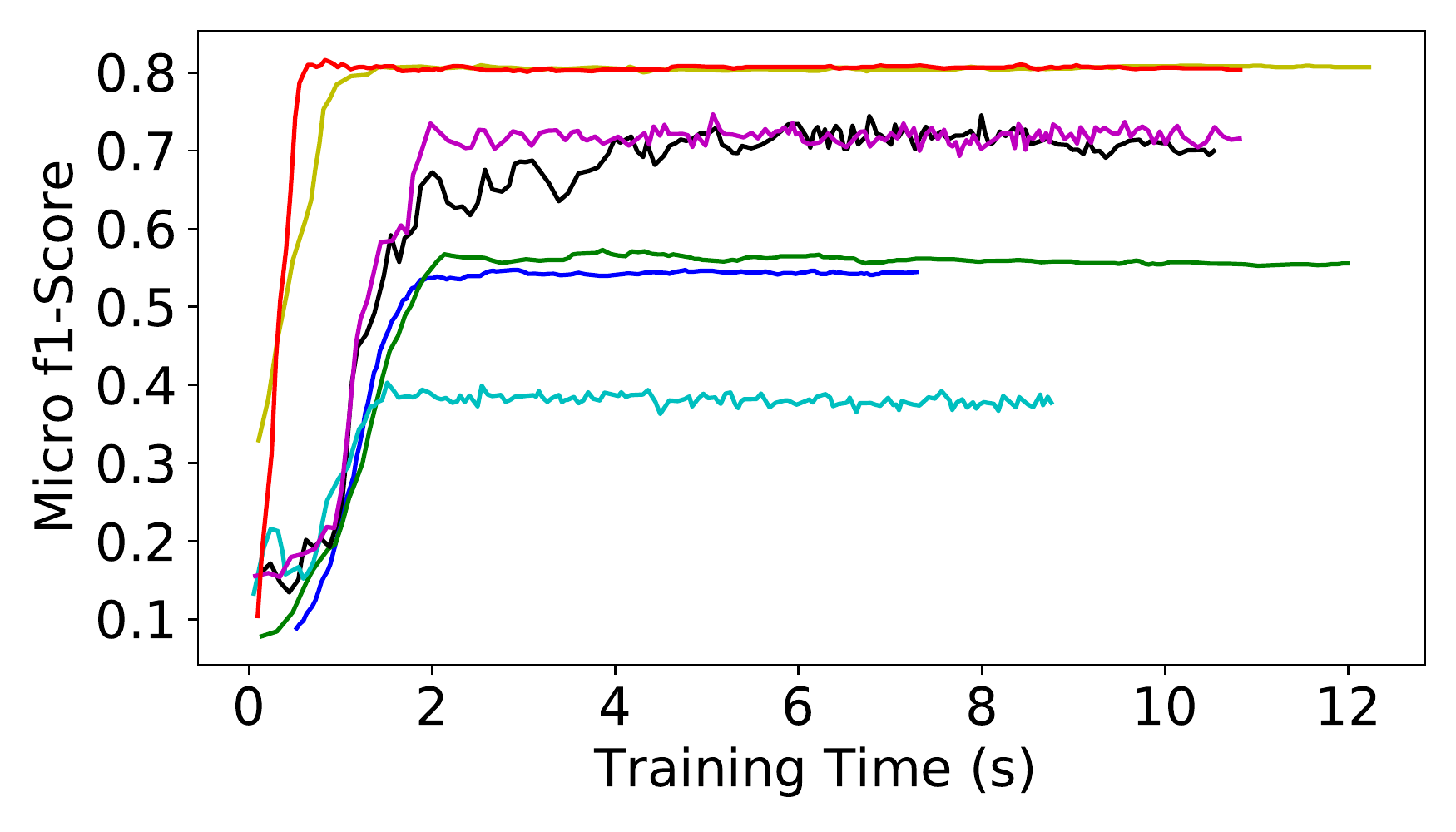}}} \hfill
\subfloat[Flicker]{{\includegraphics[scale=0.39]{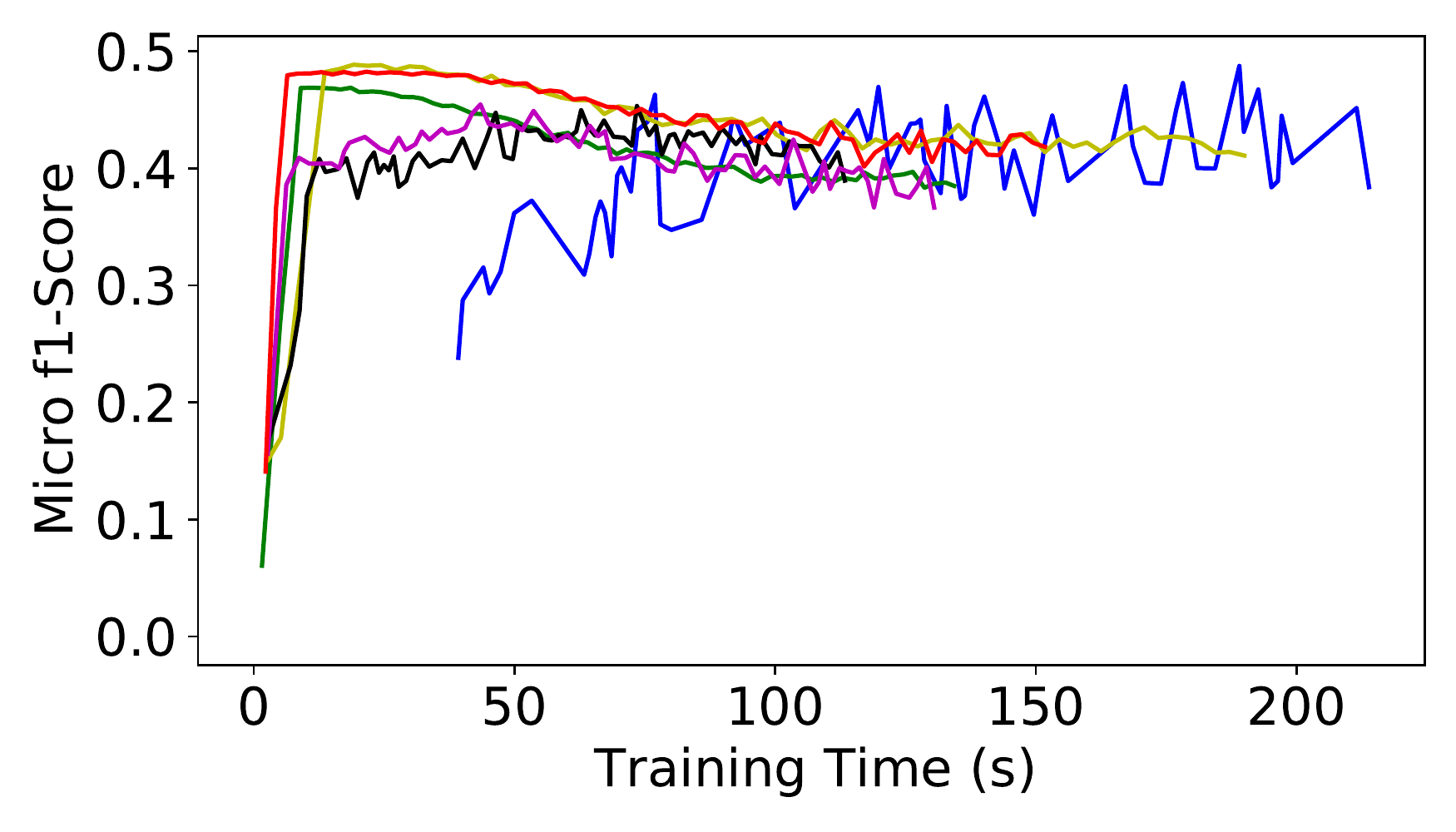}}}\hfill
\subfloat[Pubmed]{{\includegraphics[scale=0.39]{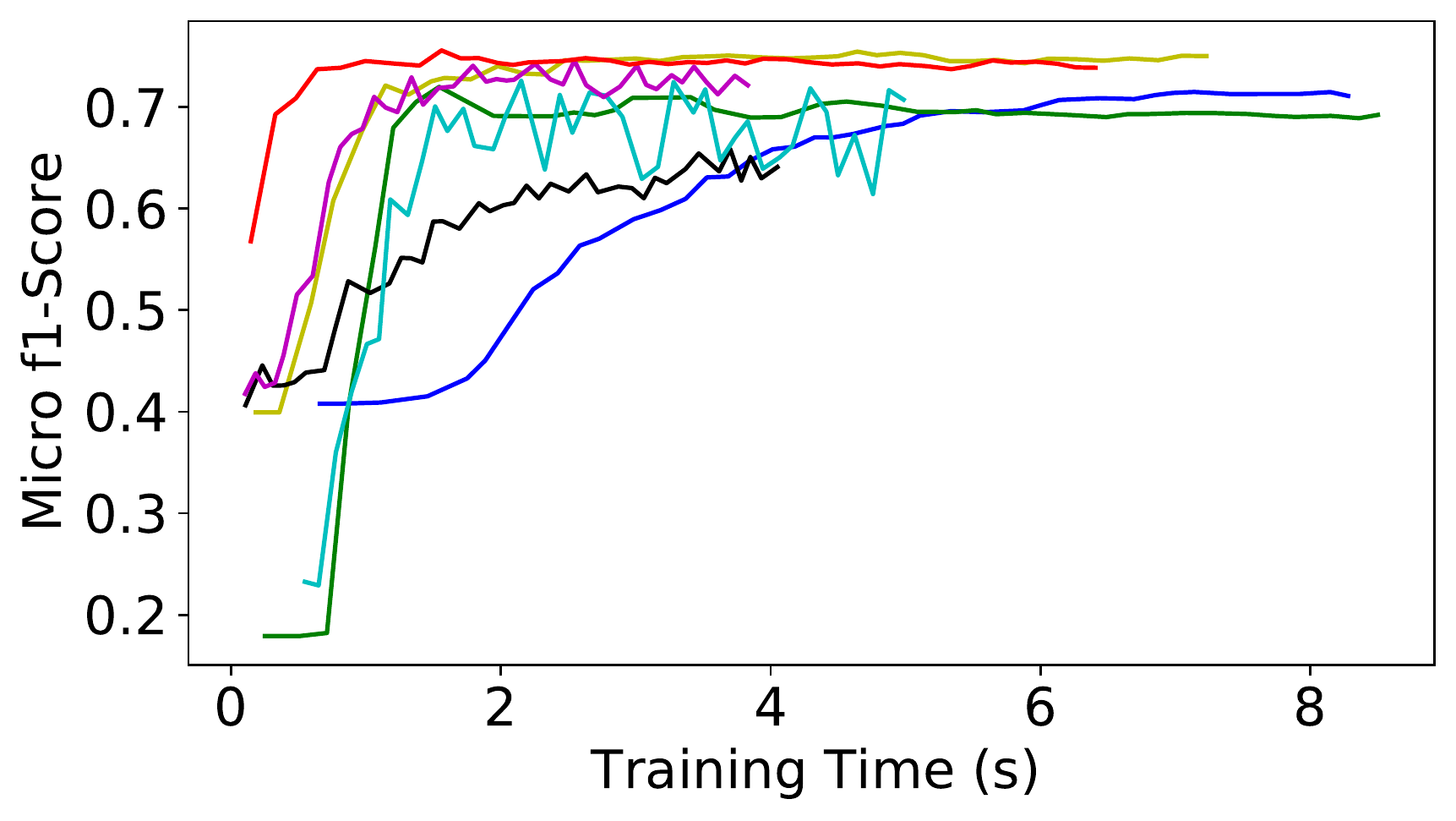}}}\hfill
\subfloat[Reddit]{{\includegraphics[scale=0.42]{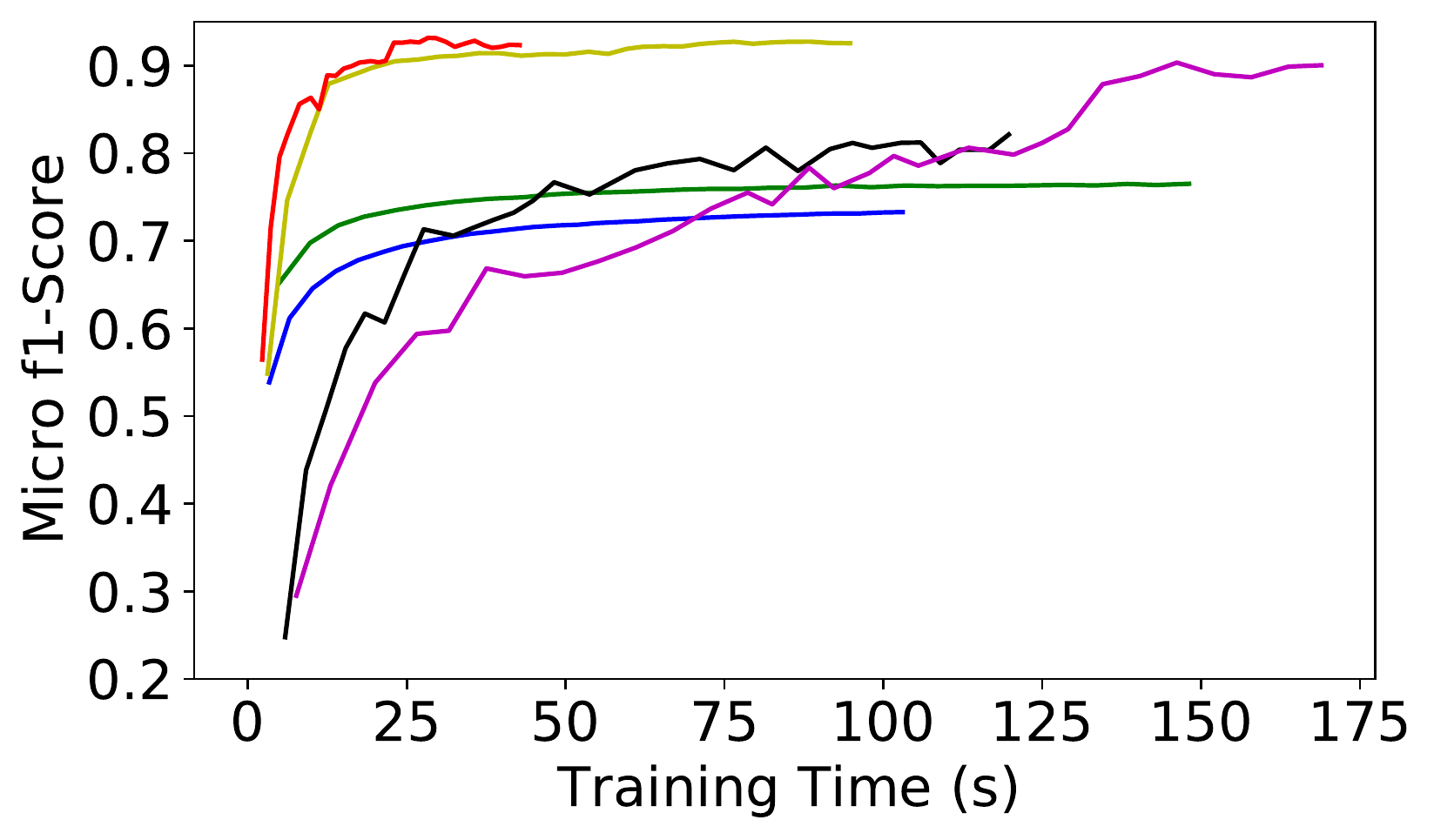}}}\hfill
\caption{The accuracy curves of test data on Cora, Flicker, Pubmed and Reddit}
\label{exp:best model accuracy}
\end{figure}

We now report the experimental results of the proposed framework and baselines under the experimental settings described above in terms of computational complexity and accuracy to analyze the effectiveness of the proposed framework. In this experiment, we have used 4 GPUs with the following parameters on all datasets. Best performing parameters for  Cora are $l=3$, $h=128$; Pubmed $l=2$, $h=256$; Flicker $l=4$, $h=128$; and Reddit  $l=3$, $h=256$. Figure \ref{exp:best model accuracy} shows the accuracy and convergence comparison of various methods on best combinations of model parameters for each dataset. We observe that the proposed framework helps reduce the computational complexity and showed significant improvement in accuracy for all datasets. Table \ref{tb:exp best model accuracy} illustrates the comparative analysis of our framework with baseline methods. It can be noticed that GAD has comparatively higher accuracy for Cora (\textbf{$\uparrow$0.67\%}), Pubmed(\textbf{$\uparrow$ 0.47\%}), Flicker (\textbf{$\uparrow$0.21\%}), Reddit (\textbf{$\uparrow$0.41\%}) in comparison to best performer.



\begin{figure}[htbp]
\centering
\includegraphics[scale=0.38]{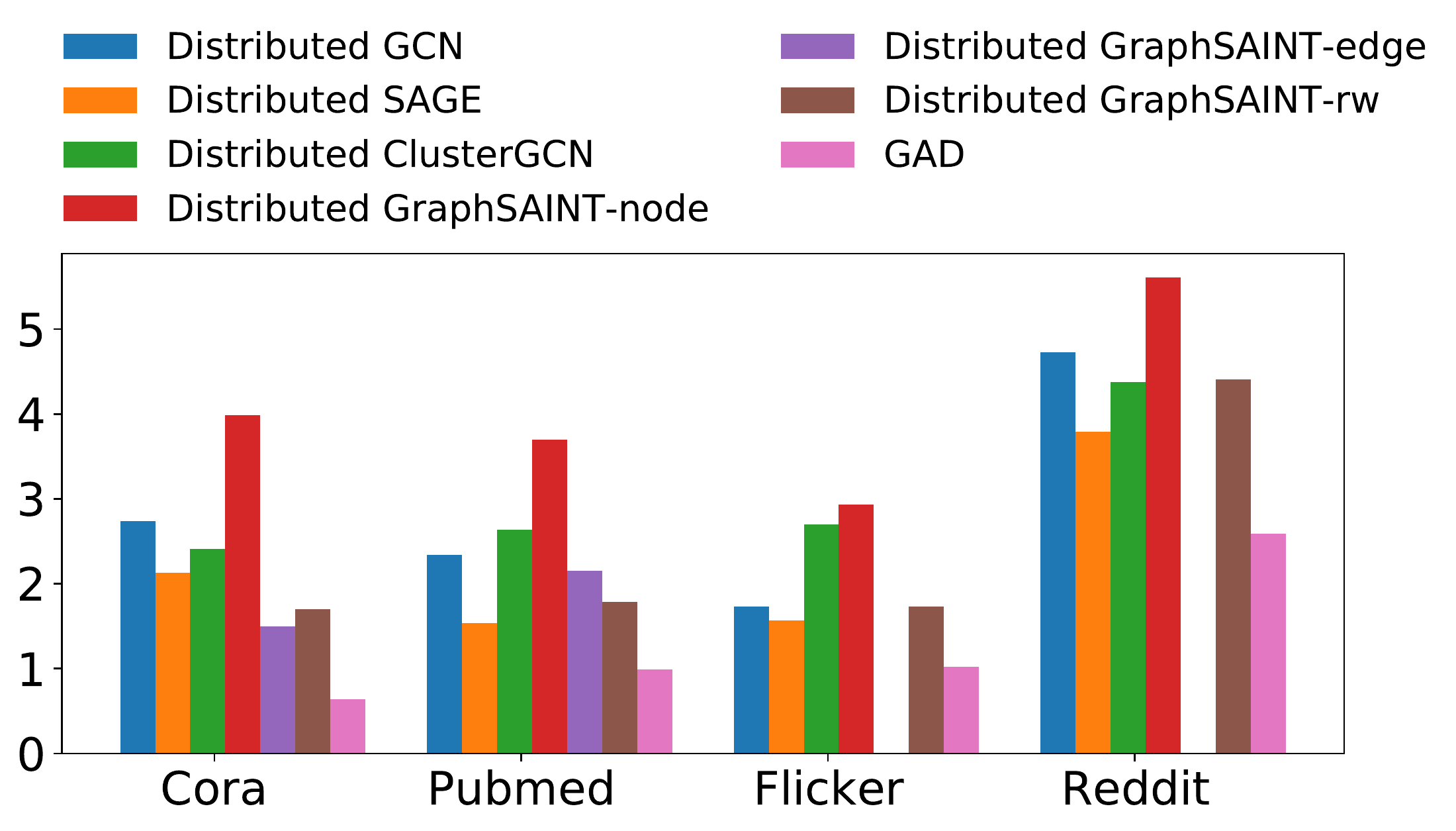}
\caption{Time cost of different GCN training methods}
\label{exp:total convergence time}
\end{figure}

In addition, the comparison of convergence time of GCN model training with benchmark methods is shown in Figure \ref{exp:total convergence time}. We average the convergence time for all datasets under the same sampling method. Results show that GAD improves the distributed training speed by 2.2x, 1.7x, 2.3x, 3.1x, 2.1x, and 1.8x compared to distributed GCN, SAGE, ClusterGCN, GraphSAINT-Node, GraphSAINT-Edge, and GraphSAINT-RW, respectively. 

Notice that our method uses the same graph partitioning method as distributed ClusterGCN \cite{DBLP:conf/kdd/ChiangLSLBH19}. Thus, we can conclude that the accuracy and training speed is improved by the graph augmentation and weighted global consensus strategies (discussed in Section \ref{method:2} and \ref{method:var measurement}). In addition, the distributed GraphSAINT-Edge sampler has higher computational complexity per epoch. Thus it does not support large-scale datasets. Thus, we did not report the comparison with distributed GraphSAINT-Edge on Flicker and Reddit datasets.

\subsection{Stability of GPU and Layers}\label{exp:Scalability of GPU and layer number}
To demonstrate the stability of deep layers in multi-GPU training, we evaluated our framework's training accuracy and training speed with the different number of layers and number of GPU.  
Generally, GCN model with more number of layers is enormously complex in training \cite{DBLP:conf/nips/HamiltonYL17, DBLP:conf/iclr/KipfW17, DBLP:conf/iclr/ChenMX18}. On the contrary, adding more layers to the network is effective and can improve the power of expressiveness which may result in improvement in performance \cite{DBLP:conf/colt/Telgarsky16}. Thus, in our framework, we aim to maintain consistent accuracy and reduce training costs as the number of GPUs increases. Table \ref{tb:layers and GPU numbers accuracy comparision} shows the GCN training accuracy with different value of $l$ and  $h$ such that $l \in \{2, 3, 4\}$ and $h \in \{128, 256, 512\}$ respectively, and different number of GPUs as 1, 2, 3, 4 for Pubmed dataset. We can observe that there is no significant loss in accuracy for single GPU

Comparing with the training accuracy by using the single machine GCN training (1 GPU) with multiple distributed GPUs under the different number of layers, there is no significant loss in accuracy i.e., the fluctuation of accuracy is less than 0.01 in the distributed GCN training environment.

\begin{figure}[htbp]
\centering
\includegraphics[scale=0.39]{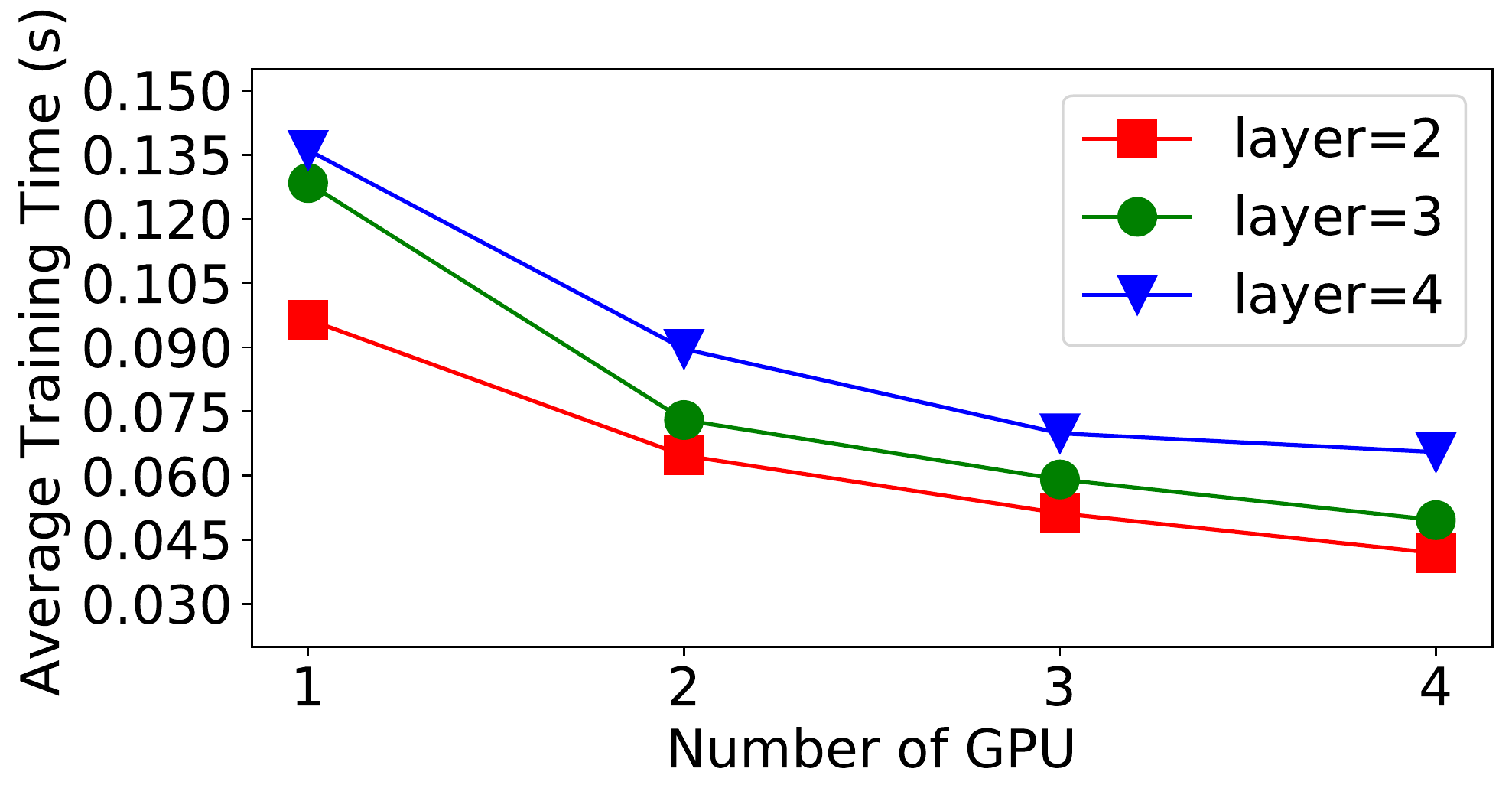}

\caption{Stability of GAD on training time when GPU and layer number vary.}
\label{fig:layer and GPU numbers time comparision}
\end{figure}

In addition, Figure \ref{fig:layer and GPU numbers time comparision} describes the average training time with variable number of GPUs (1, 2, 3, 4) and $l$ (2, 3, 4). We can notice that training time decreases with the increased in number of GPUs. 
Besides, we can observe that the time cost of GCN training decreases non-linearly. It tends to convergence into a constant with the increase of GPU number. This is because the communication and blocking take extra time during distributed training. 

\subsection{Evaluation of Graph Augmentation} \label{exp: Effects of Graph Augmentation}

\begin{table*}[htbp]
    \caption{Impact of graph augmentation when GPU number varies}
    \vspace{5pt}
    \centering
    \begin{tabular}{c c c c c c}
        \hline
        Dataset & Number of GPU & Augmentation   &   Accuracy   &   Allocated Memory per GPU (MB)   &   Communication Size (MB) \\ 
        \hline
        Cora    &   1   &   No  &   0.8067   &  63.39  &   0  \\
        Cora    &   1   &   Yes &   0.8073   &  65.12  &   0   \\
        Cora    &   4  &   No  &   0.7783   &  53.76  &    0.11 \\
        Cora    &   4   &   Yes  &   \textbf{0.8037 }  &  54.98  &   0.05   \\
        \hline
        Pubmed  &   1   &   No  &   0.7532  &  68.91  &   0   \\
        Pubmed  &   1   &   Yes &   0.7511   &  70.86  &   0   \\
        Pubmed  &   4   &   No  &   0.7138   &  39.62  &   3.61   \\
        Pubmed  &   4   &   Yes &   \textbf{0.7423 }  &  40.14  &    1.72  \\
        \hline   
    \end{tabular}
    \label{exp:comparison of subgraph augmentation}
\end{table*}



To validate the effectiveness of graph augmentation in the distributed GCN training (described in Section \ref{method:2}), we evaluated the performance in terms of accuracy, allocated memory, and communication size with and without graph augmentation in single-machine training (1 GPU) and distributed training (multi-GPUs).
Table \ref{exp:comparison of subgraph augmentation} describes the comparative analysis of the proposed framework on Cora and Pubmed datasets. We can notice that comparing with single machine training (1 GPU) without graph augmentation, the distributed training (4 GPUs) without graph augmentation has comparatively low accuracy for Cora (\textbf{$\downarrow$2.84\%}) and Pubmed (\textbf{$\downarrow$3.94\%}). This is because there is some information loss during the communication of node information in the distributed training. When the graph augmentation is applied, the accuracy is increased significantly to achieve the rate being close to the single machine training accuracy in both datasets.


In addition, we also tested the memory and communication size with/without graph augmentation when the number of GPUs is 4 for both datasets. 
We can observe a significant decrease (54.55\% and 52.35\% for Cora and Pubmed respectively) in communication overhead. This is because we store the important nodes of other processors when these nodes are frequently used and communicated to the current processors. In addition, the average allocated memory is slightly increased by 2.23\%. This is because we need to replicate some communication nodes and add them to each subgraph. Furthermore, we also find that using more GPU in a distributed environment can effectively reduce the allocated memory per GPU. It is because the training data stored in multi-GPUs is less than a single GPU.

\begin{figure}[htbp]
\centering

\subfloat[With graph augmentation]{{\includegraphics[scale=0.39]{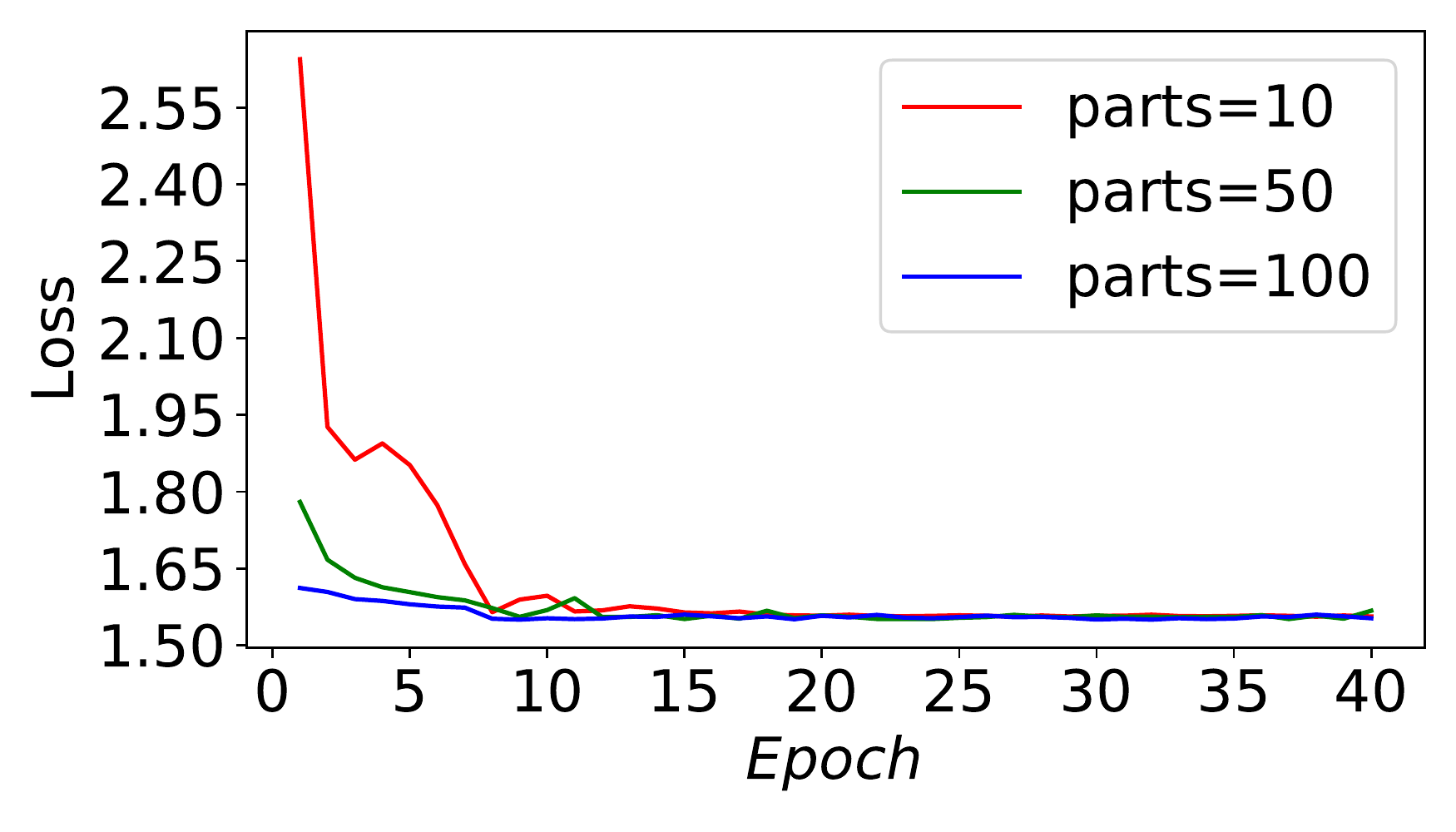}}} \hfill
\\
\subfloat[Without graph augmentation]{{\includegraphics[scale=0.39]{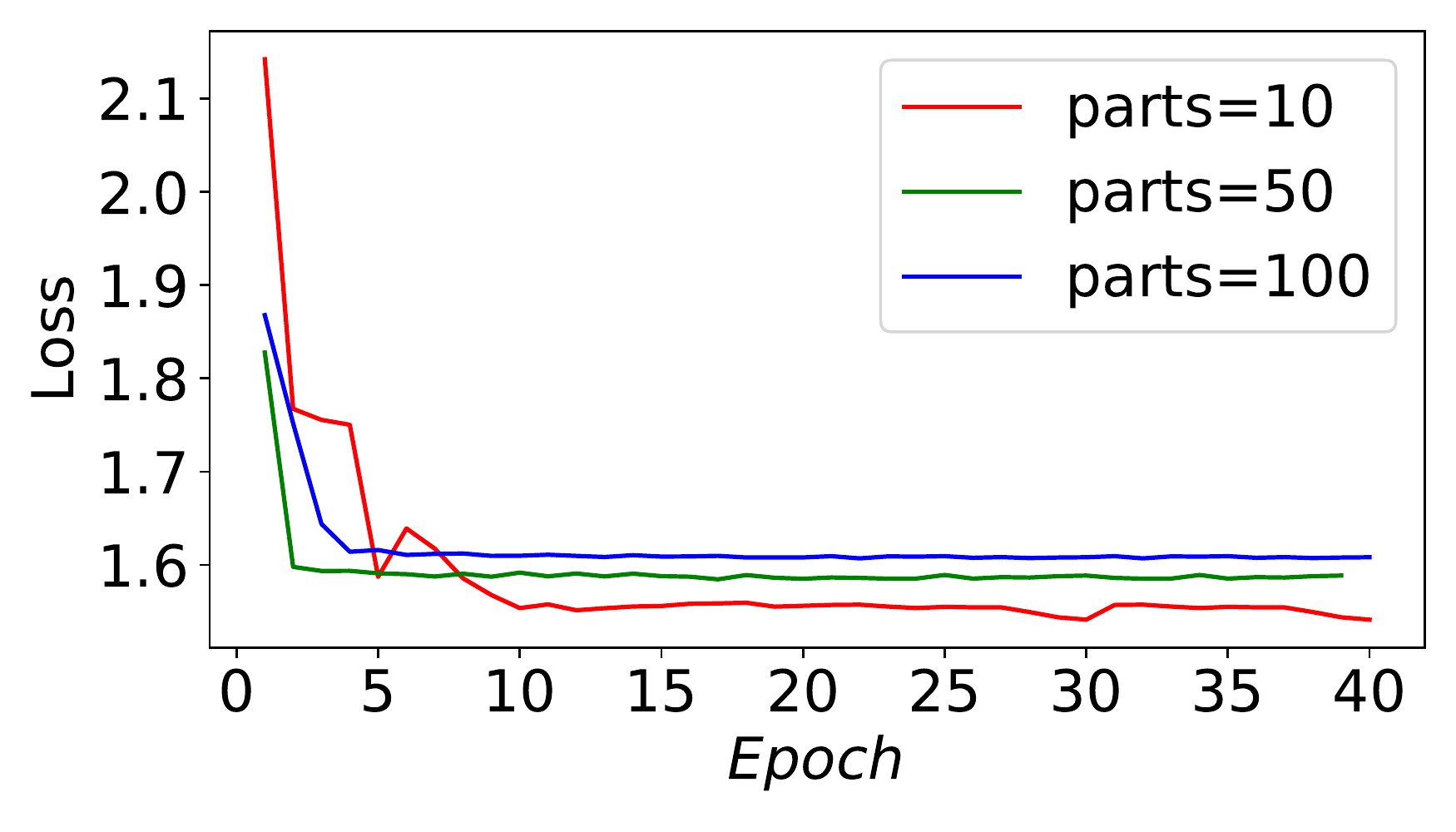}}}\hfill


\caption{Impact of graph augmentation when partition number varies}
\label{fig:partition}
\end{figure}

Furthermore, Figure \ref{fig:partition} shows the effect of graph augmentation on the number of graph partitions when the number of partitions is set as 10, 50, 100, and $l=4$, $h=512$ in Pubmed dataset. Specifically, Figure \ref{fig:partition}.a, Figure \ref{fig:partition}.b present the results of the loss convergence with/without subgraph augmentation respectively. From the results, we can see that GCN training without graph augmentation has different loss under different partition numbers. Thus, we need to balance the loss and partition numbers. When the graph augmentation is used, the convergence of loss for different partition numbers keeps consistent. This is because the information loss caused by partitions is compensated by graph augmentation. 

\subsection{Evaluation of weighted global consensus} \label{exp:Effects of weighted global consensus}

\begin{figure}[htpb]
\centering

\subfloat[Number of partition=50]{{\includegraphics[scale=0.39]{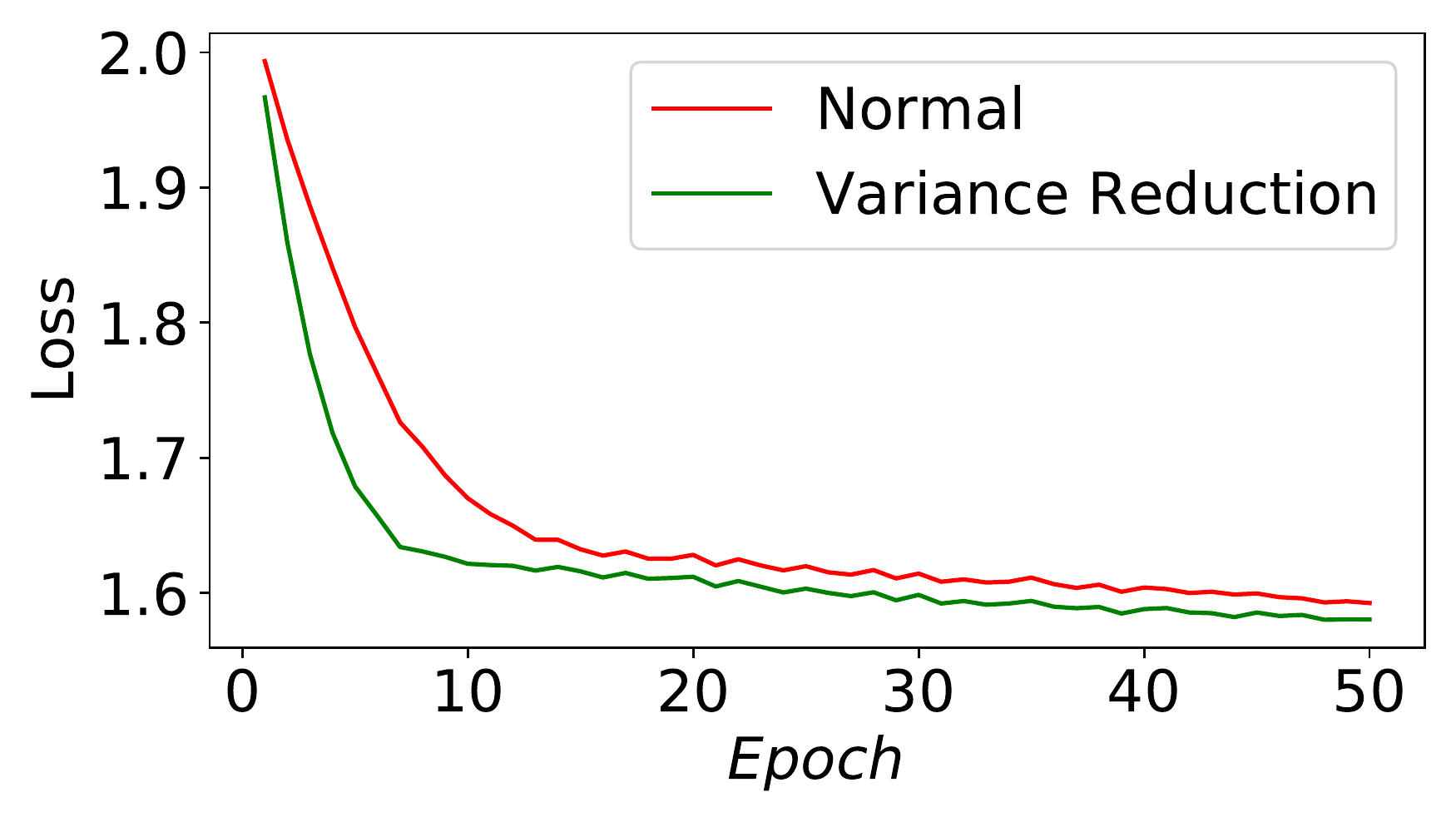}}} \hfill
\subfloat[Number of partition=100]{{\includegraphics[scale=0.39]{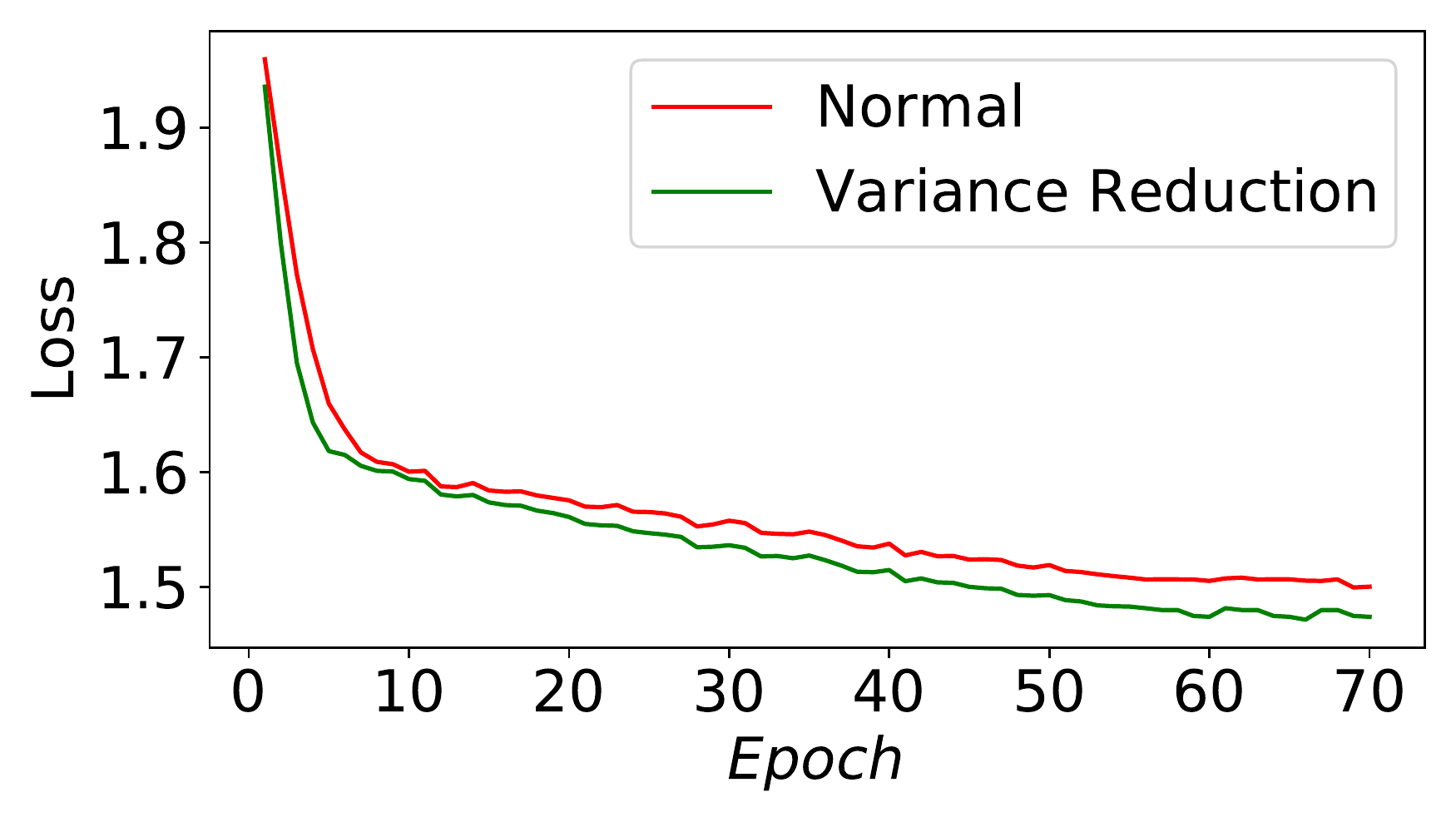}}}\hfill

\caption{Impact of weighted global consensus }
\label{fig:var_novar}
\end{figure}

To validate the effectiveness of weighted global consensus, we compared the impact of weighted global consensus on convergence speed and loss. Figure \ref{fig:var_novar}.a, Figure \ref{fig:var_novar}.b show results on the Flicker dataset when the number of partitions is set as 50, 100, respectively. In this evaluation, we set $h=128$ and $l=4$. The experimental results show lower convergence loss and faster convergence speed when the proposed weighted global consensus is utilized. This is because we have assigned a higher weight to the parameters when their corresponding subgraphs have a lower variance, which can speed up the convergence of the model towards the optimal direction. Besides, it also reduced the influence of subgraphs with high variance by decreasing the weight of their corresponding parameters. Therefore, it makes the performance being nearly-close to the optimal value.


\section{Conclusion}
In this paper, we presented a novel Graph Augmentation-based Distributed GCN training framework (GAD) to reduce the communication costs of processors and accelerate GCN training efficiency. At the same time, achieve the high accuracy of GCN training in the distributed training environment. We designed GAD-Partition to generate subgraphs with important communication information through the Monte-Carlo-based node importance measurement and the depth-first sampling strategy. In addition, we developed a GAD-Optimizer to reduce the impact of high graph variance by assigning a low importance weight to the gradient of the subgraph when GCN parameters are updated. To show the performance of our proposed framework, we conducted an extensive experiment and compared six baseline methods on four real-world datasets. The experimental results demonstrated high accuracy, good stability, and low communication costs of our GAD in different configuration settings, e.g., the accuracy of GAD can achieve about 0.816, 0.756, 0.49, and 0.931 on the test data of Cora, Pubmed, Flicker, and Reddit. Besides, there is a 1.7-3.1 times improvement in training time than baseline methods. Thus, we can conclude that the proposed Graph Augmentation-based Distributed GCN training framework can be deployed for learning large-scale graphs in real applications.

\section*{Acknowledgment}
This research is partially supported by Australian Research Council Linkage Project LP180100750.

\ifCLASSOPTIONcaptionsoff
  \newpage
\fi



%




\bibliographystyle{IEEEtran}
\bibliography{main}  

%

\begin{IEEEbiography}[{\includegraphics[width=1in,height=1.25in,clip,keepaspectratio]{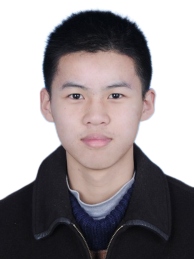}}]{Taige Zhao} received his M.S degree from The University of Sydney, Australia, in 2020. Currently, he is a PhD student in the School of Information Technology, Deakin University. His research interests include graph represents learning, distributed computing, deep learning and natural language processing.
\end{IEEEbiography}

\begin{IEEEbiography}[{\includegraphics[width=1in,height=1.25in,clip,keepaspectratio]{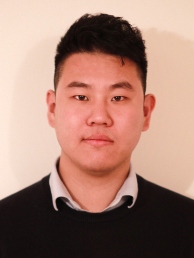}}]{Xiangyu Song} received the B.S degree from Beijing Jiaotong University, China, in 2013. Currently, he is a PhD student in the School of Information Technology, Deakin University. His research interests include educational knowledge data mining, representation Learning, and deep learning on graph.
\end{IEEEbiography}

\begin{IEEEbiography}[{\includegraphics[width=1in,height=1.25in,clip,keepaspectratio]{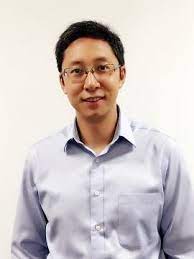}}]{Jianxin Li} (Member, IEEE) received his B. Sc. and M. Eng. degree from Northeastern University, China, in 2002 and 2015, the PhD degree in computer science from the Swinburne University of Technology, Australia, in 2009. He is an Associate Professor in Data Science at the School of Information Technology, Deakin University. His research interests include graph database query processing \& optimization, social network analytics \& computing, representation learning, and online learning analytics. 	
\end{IEEEbiography}

\begin{IEEEbiography}[{\includegraphics[width=1in,height=1.25in,clip,keepaspectratio]{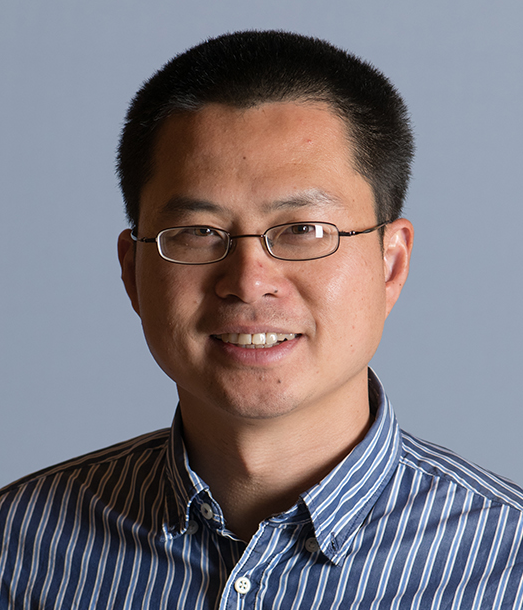}}]{Wei Luo} is a Senior Lecturer in Data Science at Deakin University. Wei's current research focuses on machine learning under uncertainties. Wei holds a PhD in computer science from Simon Fraser University, where he received training in statistics, machine learning, computational logic, and modern software development.
\end{IEEEbiography}

\begin{IEEEbiography}[{\includegraphics[width=1in,height=1.25in,clip,keepaspectratio]{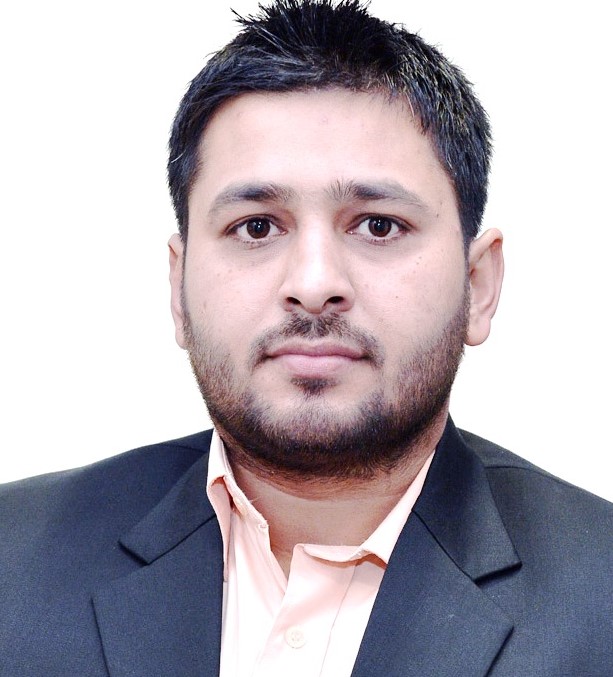}}]{Imran Razzak} is a Senior Lecturer in Computer Science at School of Information Technology, Deakin University. He received PhD in data science from University of Technology, Sydney, Master’s in Machine Learning International Islamic University. His area of includes machine learning in general with emphasis to natural language processing and image analysis to solve real world problems related to health, finance, social media, and security.  

\end{IEEEbiography}



\vfill


\end{document}